\documentclass[10pt,journal,compsoc]{IEEEtran}

\usepackage{multirow}
\usepackage{makecell}
\usepackage{multirow}
\usepackage{mathrsfs}
\usepackage[ruled]{algorithm2e}
\usepackage{cellspace}
\usepackage{amsmath}

\usepackage{amsfonts}
\usepackage{xcolor}
\usepackage{setspace}
\usepackage{makecell}
\usepackage{bm}
\usepackage{hyperref}
\usepackage{graphicx}
\usepackage{paralist}
\usepackage{enumitem}
\usepackage{amssymb}
\usepackage{multirow}
\usepackage{xcolor}
\usepackage{bm}
\usepackage{multirow}
\usepackage{longtable}

\newcommand{\tabincell}[2]{
\begin{tabular}{@{}#1@{}}#2\end{tabular}
}

\newcommand{\vpara}[1]{\vspace{0.07in}\noindent\textbf{#1 }}

\def\!#1{\bm{#1}}
\definecolor{comment}{RGB}{70, 150, 60}
\ifCLASSOPTIONcompsoc
  \usepackage[nocompress]{cite}
\else
  \usepackage{cite}
\fi

\ifCLASSINFOpdf

\else

\fi

\begin{document}

\title{Does Negative Sampling Matter? A Review with Insights into its Theory and Applications}

\author{Zhen Yang, 
        Ming Ding,
        Tinglin Huang,
        Yukuo Cen,
        Junshuai Song,
        Bin Xu$^{\star}$, 
        Yuxiao Dong, 
        and Jie Tang$^{\star}$,~\IEEEmembership{Fellow IEEE}
\IEEEcompsocitemizethanks{\IEEEcompsocthanksitem Zhen Yang and Ming Ding are with the Department of Computer Science and Technology, Tsinghua University, Beijing, 100084, China. E-mail:\{yangz21, dm18\}@mails.tsinghua.edu.cn \\ 
\IEEEcompsocthanksitem Tinglin Huang is with the Department of Computer Science, Yale University,
New Haven, CT 06511, USA.E-mail:tinglin.huang@yale.edu \\
Work was done during his visit to Tsinghua University. \\
\IEEEcompsocthanksitem Yukuo Cen is with Zhipu AI, Beijing, China, 100084. E-mail: yukuo.cen@zhipuai.cn \\
\IEEEcompsocthanksitem Junshuai Song is with the Technology and Engineering Group, Tencent, Beijing, China, 100080. E-mail:jasonjssong@tencent.com \\
\IEEEcompsocthanksitem Bin Xu, Yuxiao Dong, and Jie Tang are with the Department of Computer Science and Technology of Tsinghua University, Beijing, China, 100084.  E-mail:\{xubin, yuxiaod, jietang\}@tsinghua.edu.cn \\
$^{\star}$Corresponding authors.
}}

\markboth{Journal of IEEE TRANSACTIONS ON Pattern Analysis and Machine Intelligence,~Vol.~14, No.~8, August~2023}%
{Shell \MakeLowercase{\textit{et al.}}}

\IEEEtitleabstractindextext{%
\begin{abstract}
Negative sampling has swiftly risen to prominence as a focal point of research, with wide-ranging applications spanning machine learning, computer vision, natural language processing, data mining, and recommender systems. This growing interest raises several critical questions: Does negative sampling really matter? Is there a general framework that can incorporate all existing negative sampling methods? In what fields is it applied? Addressing these questions, we propose a general framework that leverages negative sampling.
Delving into the history of negative sampling, we trace the development of negative sampling through five evolutionary paths. We dissect and categorize the strategies used to select negative sample candidates, detailing global, local, mini-batch, hop, and memory-based approaches. Our review categorizes current negative sampling methods into five types: static, hard, GAN-based, Auxiliary-based, and In-batch methods, providing a clear structure for understanding negative sampling. Beyond detailed categorization, we highlight the application of negative sampling in various areas, offering insights into its practical benefits. Finally, we briefly discuss open problems and future directions for negative sampling.  
\end{abstract}

\begin{IEEEkeywords}
Negative Sampling Framework; Negative Sampling Algorithms; Negative Sampling Applications
\end{IEEEkeywords}}

\maketitle

\IEEEdisplaynontitleabstractindextext

\IEEEpeerreviewmaketitle

\label{sec:introduction}
\section{Introduction} 

\textbf{Does negative sampling matter?} Negative sampling (NS) is a critical technique used in machine learning, designed to enhance the efficiency of models by selecting a small subset of negative samples from a vast pool of possible negative samples (i.e., non-positive samples). A prime example of NS in action is found in the word2vec~\cite{mikolov2013distributed}, particularly within its Skip-Gram architecture. The Skip-Gram model is designed to predict the likelihood of occurrence of nearby context words given a target word within a sequence. Formally, for a given sequence of training words $w_1,w_2,\cdots, w_T$, the objective is to maximize the average log probability of context words within a specified window on either side of the target word, which can be presented as:
\begin{equation*}
    \min -\frac{1}{T} \sum_{t=1}^{T} \sum_{-c \leq j \leq c, j\neq 0} \log p(w_{t+j}|w_t)
\end{equation*}
where $T$ is the length of the word sequence, $c$ is the size of the training context (that is a window of words around the target word), $w_t$ is the target word, $w_{t+j}$ are the context words within the window around $w_t$.

The probability $p(w_{t+j}|w_t)$ represents the likelihood of encountering a context word $w_{t+j}$ given a target word $w_t$, defined by the softmax function:
\begin{equation*}
    p(w_{t+j}|w_{t}) = \frac{exp({v'_{w_{t+j}}}^{T} v_{w_{t}})}{\sum_{w=1}^{W}exp({v'_{w}}^{T} v_{w_{t}})}
\end{equation*}
where $v_w$ and $v'_w$ represent the ``input'' and ``output'' vector embeddings of a word $w$, and $W$ denotes the total number of words in the vocabulary. 

The objective of the training process is to optimize these word vector embeddings to maximize the probability for true context words while minimizing it for all other words. However, implementing the Skip-Gram model with a standard softmax function poses significant computational challenges, particularly with large vocabularies. Each training step requires calculating and normalizing the probabilities of all words in the vocabulary relative to a given target word, which is computationally intensive. 

Negative sampling is leveraged to simplify this. Rather than considering all words in the vocabulary, the model only considers a small subset of negative words (not present in the current context) along with the actual positive context words. This substantially reduces the computational burden. Besides, negative sampling changes the training objective. Instead of predicting the probability distribution across the entire vocabulary for a given input word, it redefines the problem as a binary classification task. For each pair of words, the model predicts whether they are likely to appear in the same context (positive samples) or not (negative samples). The objective of the Skip-Gram model with negative sampling can be represented as:
    \begin{equation*}
       \min -\log \sigma({v'_{w_{t+j}}}^{T} v_{w_t}) - \sum_{i=1}^{k} \mathbb{E}_{w_i \sim p_n(w)} [\log \sigma(-{v'_{w_{i}}}^{T} v_{w_t})]
    \end{equation*}
where $k$ is the number of negative samples, $w_i$ are negative words sampled for the target word $w_t$, $\sigma(\cdot)$ is the sigmoid function and $p_n(w)$ is the negative sampling distribution that is designed as the 3/4 power of word frequencies.

This example demonstrates the importance of negative sampling.
Without it, the model would need to compute the co-occurrence probability with every other word for each training instance, which is computationally expensive, particularly for large vocabularies. By employing negative sampling, word2vec significantly reduces the computational burden and achieves great performance.

In a broader perspective, negative sampling is a critical technique employed across a variety of fields, such as recommendation systems (RS)~\cite{rendle2012bpr}, graph representation learning (GRL)~\cite{perozzi2014deepwalk,tang2015line,grover2016node2vec,yang2020understanding,huang2021mixgcf}, knowledge graph learning (KGE)~\cite{krompass2015type, cai2017kbgan}, natural language processing (NLP)~\cite{mikolov2013distributed,grbovic2015context,zhang2018gneg,wu2021esimcse}, and computer vision (CV)~\cite{schroff2015facenet, wu2017sampling, chuang2020debiased, kalantidis2020hard, robinson2020contrastive, wu2020conditional}. The core principle of negative sampling -- selecting a representative subset of negative samples to improve the efficiency and effectiveness of the learning process -- remains consistent across different domains. 

\begin{figure}[t]
  \centering  \includegraphics[width=0.45\textwidth]{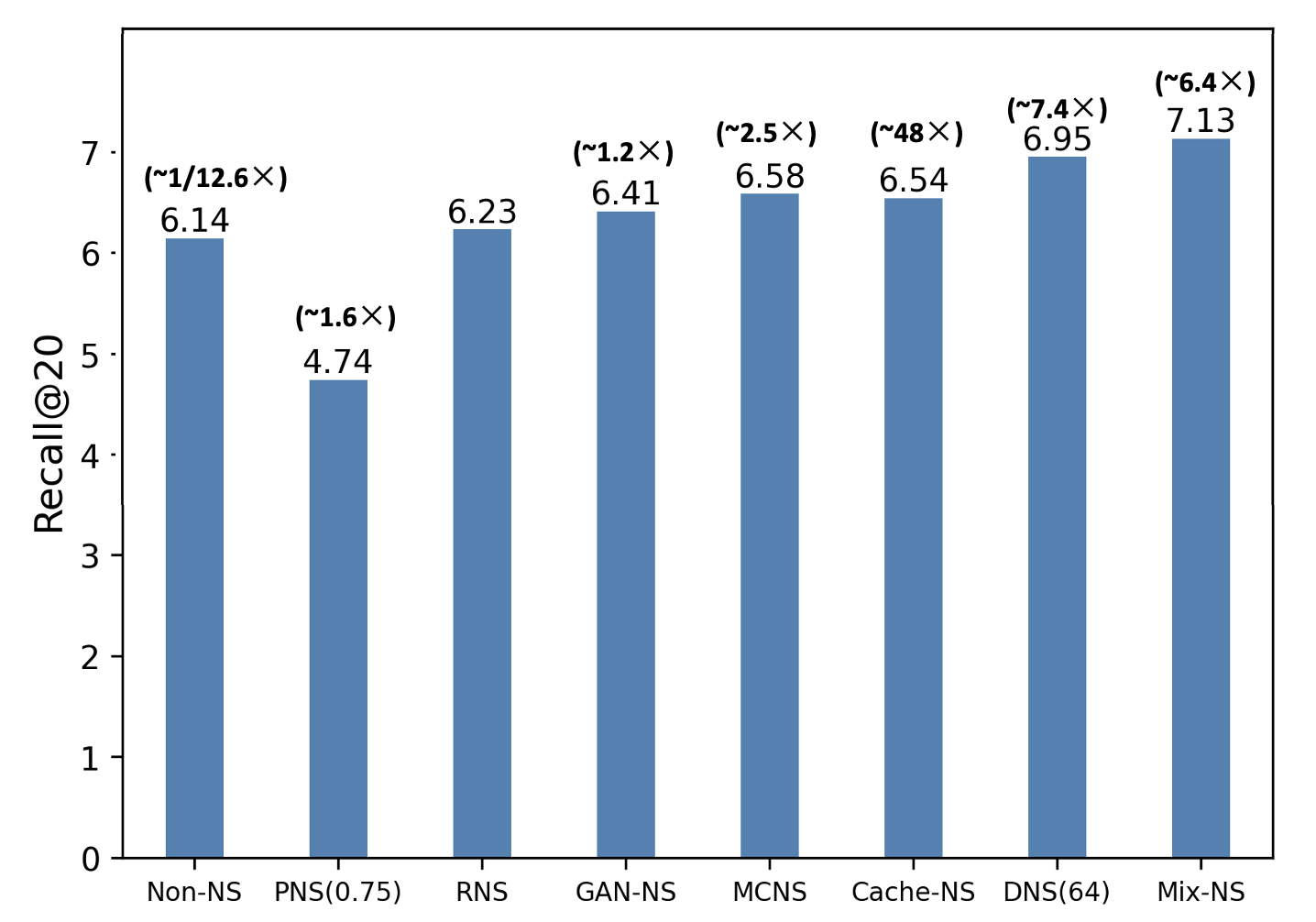}
  \vspace{-0.3cm}
  \caption{Performance and Convergence speed comparison in the RS domain with the Non-NS and other methods using negative sampling on the Yelp2018 datasets with LightGCN as an encoder. PNS($\alpha$) denotes popularity-based negative sampling with various power parameters $\alpha$. DNS($\beta$) represents dynamic negative sampling with a different number of negative sample candidates $\beta$. The details of the negative sampling methods can be found in Section~\ref{sec:trace_back}.}
  \label{fig: ns_performance}
  \vspace{-0.4cm}
\end{figure}

To clearly verify the improvements, we compare a basic method with other NS methods on performance and convergence speed in recommendation system. As shown in Figure~\ref{fig: ns_performance}, compared with the basic RNS, the method without negative sampling (Non-NS) achieves a close performance but needs a longer training time since it uses all negatives for model training. Compared to RNS, other advanced NS methods can reduce the computational burden, accelerate training convergence(\textbf{up to $\sim 48 \times$} with Cache-NS), degrade performance bias, and boost performance (\textbf{$14\%$ improvement} with Mix-NS). Compared with RNS, PNS (0.75) achieves a faster convergence of $1.6 \times$ but reduces performance by $24\%$. Compared with Mix-NS, DNS (64) shows faster convergence but slightly degrades performance.

\begin{figure*}[tbp]
  \centering
  \includegraphics[width=0.98\textwidth]{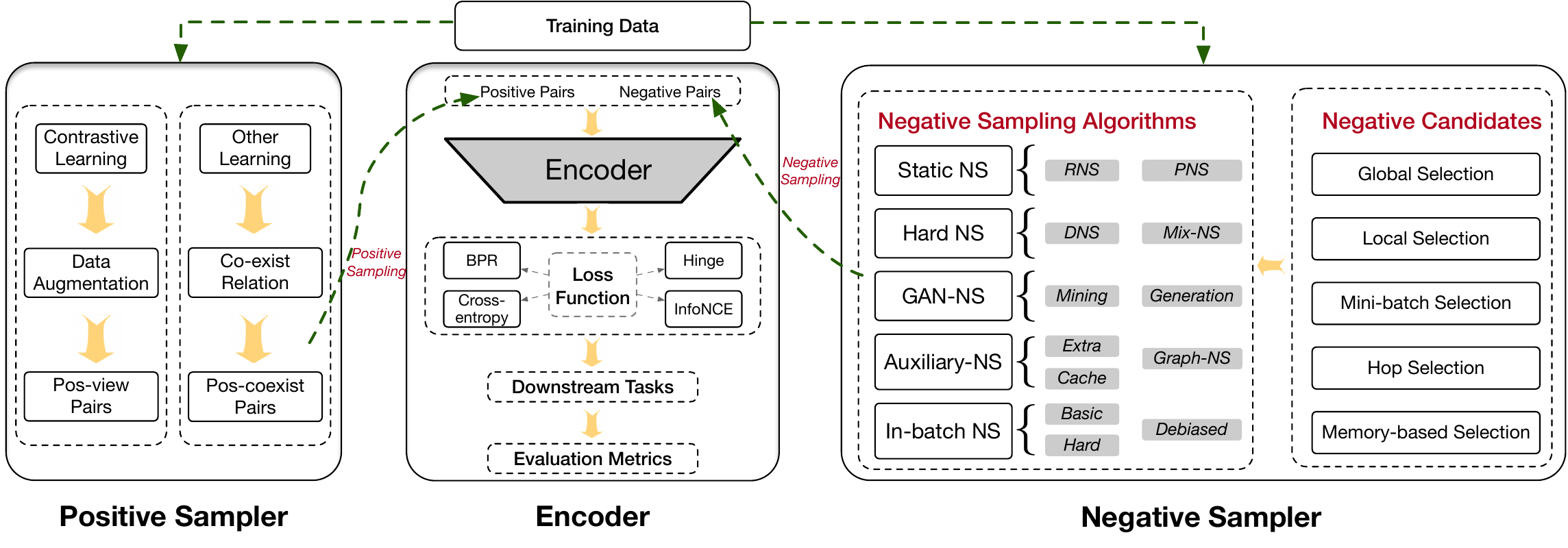}
  \vspace{-0.4cm}
  \caption{An illustration of a general framework that uses negative sampling. Positive and negative pairs are sampled implicitly or explicitly by positive and negative samplers respectively, both of them composing the training data. An encoder is applied for latent representation learning in various domains. In contrastive learning, positive pairs (i.e., Pos-view Pairs) are derived from data augmentations of the same instance or different perspectives of the same entity, while in other domains, such as metric learning, positive pairs (Pos-coexist Pairs) are the other instances in the dataset. }  
  \label{fig: ns_framework}
  \vspace{-0.3cm}
\end{figure*}

Despite the widespread application of NS, its implementation can differ greatly across domains. This raises the question: Is there a general framework that can incorporate all the NS methods and be applicable to different domains? Our survey aims to answer this by formalizing negative sampling and introducing a general framework tailored for model training (See Figure~\ref{fig: ns_framework}). Under this framework, we categorize different NS methods to make them more accessible and comparable. Besides, we summarize three critical aspects to be taken into account when designing a better NS method: (1) where to sample from? (2) how to sample? and (3) in which field should it be applied?

\vpara{Contributions.} The main contributions of this survey:

\begin{itemize}[leftmargin=*]
    \item We highlight the importance of negative sampling and propose a general framework that can incorporate existing NS methods from various domains. 

    \item We identify three aspects that should be considered for designing negative sampling: negative sample candidates, negative sampling distributions, and negative sampling applications. We sum up five selection methods for negative sample candidates and categorize existing negative sampling methods into five groups.
    
    \item We report the characteristics of negative sampling methods in various domains and demonstrate the pros and cons of each type of negative sampling method. We summarize several open problems and discuss the future directions for negative sampling.

\end{itemize}

\noindent \textbf{Survey Organization.}
The rest of this survey is organized as follows. In Section~\ref{sec:trace_back}, we provide the overview of negative sampling, first briefly tracing back the development history and then giving a general framework. Section~\ref{sec:algorithms} reviews existing NS algorithms, as well as the pros and cons of each category. Section~\ref{sec:application} further explores applications of NS in various domains. Finally, we discuss the open problems, future directions, and our conclusions in Section~\ref{sec:discussion} and Section~\ref{sec:conclusion}.

\noindent \textbf{Variables Definitions.}
Here, we elucidate the meanings of the variables employed in our survey. $x$, $x^+$, and $x^-$ denote an anchor sample, a positive sample, and the selected negative sample respectively. $p_n$ represents a designed negative sampling distribution. $x'$ denotes a negative candidate within the pool of negative sample candidates $\mathcal{C}$. These candidates are the potential selections for the negative sample $x^-$, as determined by the negative sampling strategy. $S(\cdot)$ is a function used to measure the similarity between samples, such as dot product~\cite{zhang2013optimizing,yang2020understanding}, L1 and L2 norms~\cite{bordes2013translating,kipf2019contrastive}, depending on the specific requirements of the learning task. $f(\cdot)$ represents the model that maps these input samples into their respective embeddings.

\label{sec:trace}

\section{Negative Sampling}\label{sec:trace_back}
In this section, we provide a comprehensive overview of negative sampling, detailing its formalization and framework that uses negative sampling for model training, tracing its historical development from five development lines, and elaborating its important role in machine learning.

\subsection{Formalization and Framework}\label{sec:framework}
Negative sampling aims to select a subset of negative samples from a larger pool, which is a technique used to improve the efficiency and effectiveness of training in machine learning models. The formalization of negative sampling can be encapsulated in a general loss function, which is designed to maximize the similarity of positive pairs and minimize it between negative pairs. For an anchor sample $x$ and a specific positive sample $x^+$, negative sampling selects negative sample $x^-$ from a candidate pool $\mathcal{C}$ based on a negative sampling distribution $p_n(x^-)$ for model training. 
\begin{equation*}
    \mathcal{L}={l}(x,x^+,x^-), x^-\sim p_n(x^-)
\end{equation*}
where $l(\cdot)$ is a specific loss function in various domains.  

In different application domains, the specific loss function may vary based on various tasks and datasets.
\begin{itemize}[leftmargin=*]
    \item \textbf{Bayesian Personalized Ranking (BPR) Loss for Pair-based Sampling.} For a pair of positive and negative samples (pos, neg), BPR loss aims to ensure that the log-likelihood of a positive sample surpasses that of a negative sample.

    \item \textbf{Hinge Loss for Triplet-based Sampling.} For a triplet of an anchor, a positive sample, and a negative sample (anchor, pos, neg), hinge loss seeks to maintain a minimum margin between the similarity of an anchor-positive pair and that of an anchor-negative pair.
    
    \item \textbf{Cross-Entropy Loss for Single Positive with Multiple Negatives.} In cases with one positive and multiple negatives, cross-entropy loss discriminates the positive sample from the negatives. 

    \item \textbf{InfoNCE Loss in Contrastive Learning.} InfoNCE loss, common in contrastive learning frameworks, is designed to contrast positive pairs against negative pairs.
\end{itemize}

The core principle underlying the various loss functions associated with negative sampling is to enhance the model's ability to distinguish between positive and negative pairs, effectively pulling positive samples closer and pushing negative samples apart in the representation space. Detailed formulations of these functions can be found in Table~\ref{tab:ns_loss}. Based on the formalization of negative sampling, we propose a general framework that uses negative sampling for model training, which contains a positive sampler, a negative sampler, and a trainable encoder. The overall framework is illustrated in Figure ~\ref{fig: ns_framework}. The positive sampler is applied to generate positive training pairs. For example, positive pairs in recommendation are sampled explicitly from the observed user-item interactions, while these in contrastive learning are generated implicitly by data augmentations. Negative training pairs are sampled by different NS strategies via a negative sampler. 
The sampled pairs are then processed by a trainable encoder, which is tailored to the specific application domain. The trainable encoder could be graph neural networks (GNNs)~\cite{kipf2016semi, velivckovic2017graph, hamilton2017inductive} in graph representation learning, ResNet~\cite{he2016deep} in unsupervised visual representation learning, BERT~\cite{devlin2018bert} and RoBERTa~\cite{liu2019roberta} in unsupervised sentence embedding learning, Skip-Gram~\cite{mikolov2013distributed} in word embedding, TransE~\cite{bordes2013translating} and TransH~\cite{wang2014knowledge} in knowledge graph embedding.

\begin{figure*}[t]
  \centering
  \includegraphics[width=0.98\textwidth]{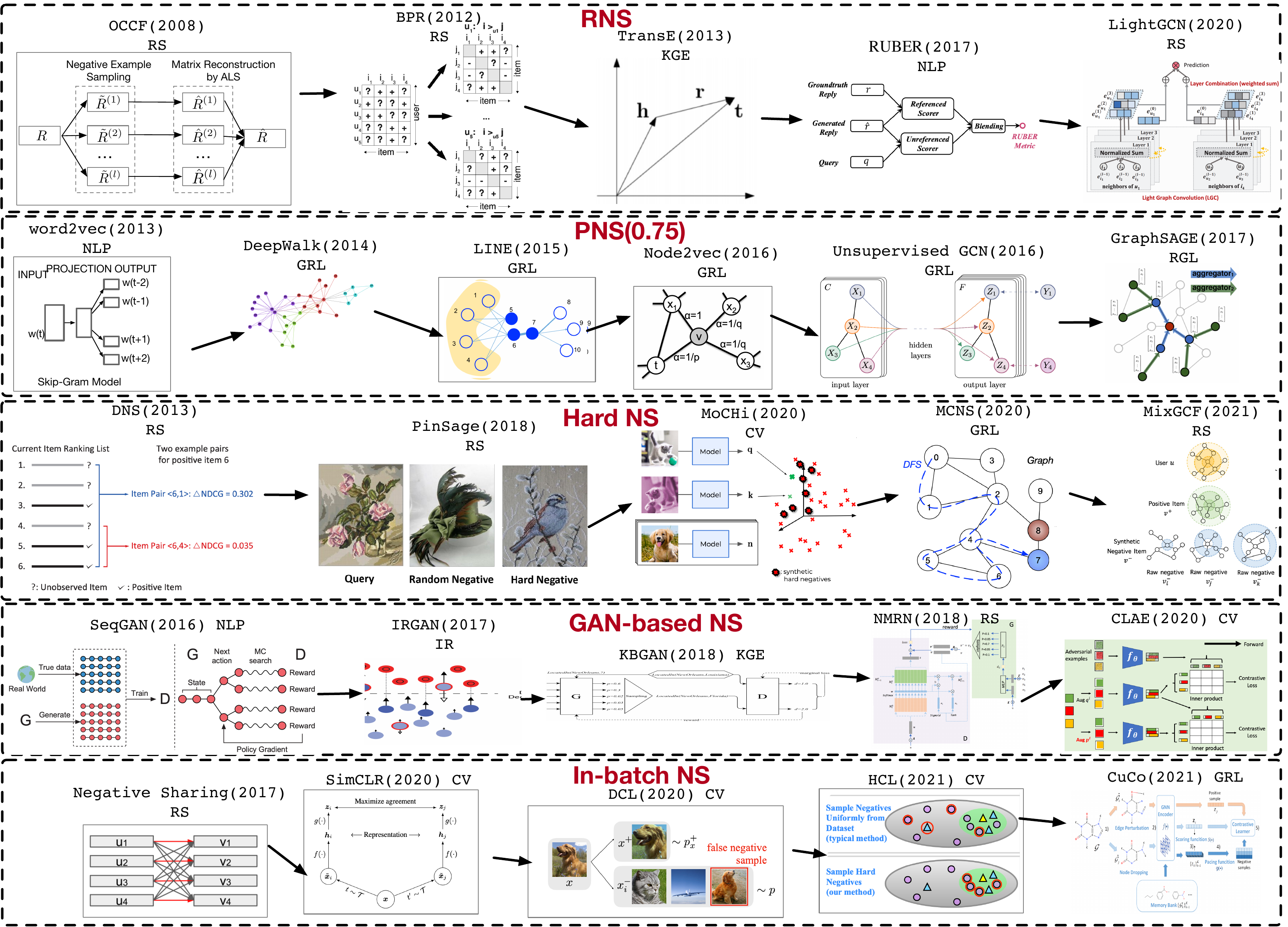}
  \vspace{-0.3cm}
  \caption{The five development lines of negative sampling. Each development line addresses different challenges and has been adopted and adapted to suit the specific needs of individual domains, such as RS, NLP, GRL, KGE, and CV.}  
  \label{fig: trace_back}
  \vspace{-0.4cm}
\end{figure*}

\begin{table}[t!]
\centering
\small
\vspace{-0.3cm}
\caption{An overview of loss functions used in a general framework that uses negative sampling for model training. $\tau$ represents a temperature parameter that scales the similarity scores. $B$ denotes the batch size.}
\renewcommand{\arraystretch}{1.5}
\setlength{\tabcolsep}{0.5mm}{
\begin{tabular}{c}  
\hline
\multirow{5}{*}{} Loss   Formulation   \\
\hline
 $\mathcal{L}_{\text{BPR}} = \text{In}\sigma( f(x^+) - f(x^-))$ \\
 $\mathcal{L}_{\text{Hinge}} = \text{max}(0, S(f(x), f(x^+)) - S(f(x), f(x^-))+\gamma) $ \\
 $\mathcal{L}_{\text{CE}} = - [\log\sigma(S(f(x), f(x^+))) + \log(\sigma(-S(f(x), f(x^-)))]$ \\
 $\mathcal{L}_{\text{InfoNCE}} = -\log \frac{exp({S(f(x), f(x^+))}/\tau)}{exp({S(f(x), f(x^+))/\tau)} + \sum_{k=1}^{B-1} exp({S(f(x), f(x_k^-))}/\tau) } $ \\
    \hline
    \end{tabular}}
\label{tab:ns_loss}
\vspace{-0.4cm}
\end{table}

\subsection{Negative Sampling History}
The history of negative sampling is a fascinating development line through the evolution of machine learning techniques. Here, we delve into the historical development of negative sampling, highlighting the motivations behind various negative sampling methods. As shown in Figure~\ref{fig: trace_back}, the earliest implementation of negative sampling is random negative sampling (RNS). The basic idea of RNS is to randomly select a subset of negative samples from the large pool, thereby reducing the computational requirements of the training process. Notably in 2008, Pan et al.~\cite{pan2008one} utilized RNS to prevent the model from developing a bias towards the majority of negative samples in one-class collaborative filtering (OCCF) for recommender systems, which effectively balanced the data. Rendle et al.~\cite{rendle2012bpr} proposed a pair-wise learning method, which leveraged pair-wise loss and RNS to solve the lack of negative data in implicit feedback. Such a RNS method is also applied to graph embedding~\cite{bordes2013translating, xiao2015transg} and GNN-based recommendation~\cite{wang2019neural, he2020lightgcn}.

Over time, as machine learning datasets expanded and model complexities increased, the limitations of RNS became evident. These randomly selected negative samples might not always be relevant for the model training. This realization leads to the development of popularity-based negative sampling (PNS)~\cite{mikolov2013distributed, perozzi2014deepwalk, grover2016node2vec, tang2015line, hamilton2017inductive}, which selects negative samples based on their occurrence frequency, popularity, or connectivity within the dataset (i.e., degree).   
This method, predicated on the hypothesis that frequently occurring negatives are more likely to represent true negatives, attempts to refine the relevance of negative samples to the dataset. The adoption of a distribution proportional to the 3/4 power of sample frequency became a hallmark of PNS, proving its efficacy in domains such as NLP~\cite{mikolov2013distributed} and graph embedding learning~\cite{perozzi2014deepwalk,grover2016node2vec,hamilton2017inductive}.

Nonetheless, both RNS and PNS exhibit a static nature, failing to adapt as model training, which can lead to suboptimal performance. This challenge paves the way for 
hard negative sampling strategies (Hard NS)~\cite{dalal2005histograms,felzenszwalb2009object,malisiewicz2011ensemble,zhang2013optimizing}, a strategy designed to identify and leverage negatives that are difficult for the model to distinguish accurately. These hard negatives, providing a more effective learning signal, align the negative sampling more closely with the model's current state, enhancing training efficiency and effectiveness. Additionally, these hard negatives provide more information for gradients in model optimization, which can accelerate convergence and boost performance. Hard NS is applied to word embedding~\cite{chen2018improving}, answer selection~\cite{rao2016noise}, knowledge graph embedding~\cite{sun2018bootstrapping}, graph embedding learning~\cite{yang2020understanding}, GNN-based recommendation~\cite{ying2018graph, huang2021mixgcf, wang2020reinforced}, and object detection~\cite{shrivastava2016training}.

The inspiration drawn from Generative Adversarial Networks (GANs)~\cite{goodfellow2014generative} leads to the development of GAN-based negative sampling (GAN-based NS)~\cite{wang2017irgan, cai2017kbgan, wang2019learning, wang2018neural, chae2018cfgan, park2019adversarial}, a technique where adversarial training and adversarial examples are leveraged to generate negative samples that closely mirror the representations of true negative samples not easily found within the dataset. In this framework, a generator serves as a negative sampler while the training model acts as the discriminator. Such methods rely on a generator that adaptively approximates the underlying negative sampling distribution. GAN-based NS is a general strategy that can be applied to recommendation, graph learning, knowledge graph learning, and unsupervised visual representation learning.

Recently, contrastive learning, particularly within unsupervised visual representation learning, adopts in-batch negative sampling (In-batch NS) as a strategy to capitalize on the efficiency of using other samples within the same batch as negatives. This approach obviates the need for external steps to generate or select negatives, accelerating the training process. For example, SimCLR~\cite{chen2020simple} utilizes other samples in the current mini-batch as negatives. MoCo~\cite{he2020momentum} maintains a memory bank to store the past several batches as negatives for model training, thereby enriching the pool of negatives. However, the effectiveness of In-batch NS is constrained by the batch size, where smaller batches might not offer a diverse array of negatives, potentially limiting the model's learning capacity. To address this, several works~\cite{kalantidis2020hard, robinson2020contrastive, wu2020conditional, hu2021adco, xie2020delving, yang2023batchsampler} have focused on introducing hard negatives into the contrastive learning paradigm.

\subsection{The Importance of Negative Sampling}

Here, we explicitly elaborate on the \textbf{importance} of negative sampling from three aspects.  
\begin{enumerate}[leftmargin=*]
    \item \textbf{Computational Efficiency.} By selecting a representative subset of negative samples, negative sampling eliminates the need for the model to consider all possible negative samples during training. Without negative sampling, the model computes the probability distribution for a sample against the entire dataset, which is computationally intensive, especially for large datasets~\cite{pan2008one,mikolov2013distributed}. Negative sampling simplifies this by transforming the task into a binary classification problem, which allows the model to differentiate between positive samples and a small number of selected negative samples. Thus, negative sampling significantly reduces the computational burden and accelerates the training process.

    \item  \textbf{Handing Class Imbalance.} Real-world datasets often exhibit a significant imbalance between positive and negative samples, resulting in performance bias. Negative sampling addresses this issue by curating a more balanced training dataset, selecting a representative subset of negatives. This method prevents the model from developing a bias towards the majority class, thereby improving the model's ability to accurately predict less frequent classes. In cases like one-class application scenarios, only target positive samples with an inherent absence of explicit negative feedback, often indiscriminately treat all non-positive samples as negatives. By judiciously choosing negatives, negative sampling ensures more equitable data for the model training. Thus, negative sampling is a key technique in addressing performance bias in models dealing with real-world data.

    \item \textbf{Improved Model Performance.} At the heart of negative sampling's value is its ability to improve model performance through the selection of negatives, particularly those that are closely aligned with positive samples in embedding space.  By focusing on a subset of more informative negative samples, the model can better capture the subtle distinctions between different samples. As highlighted in studies~\cite{cai2020all, yang2020understanding, caselles2018word2vec, wu2017sampling, schroff2015facenet}, negative samples, particularly more informative or ``hard'' negatives, contribute significantly to the gradients during the training process. \textbf{Hard negatives} refer to samples that closely resemble positive samples in feature space, making the difficult for the model to distinguish from the positives. Training with these hard negative samples forces the model to learn finer distinctions because these negatives contribute more significantly to the gradients, leading to a more effective optimization process and improvement in the model's ability to distinguish between positive and negative samples. Thus, negative sampling, especially when it involves the strategic use of hard negatives, is an essential technique for improving the performance of machine learning models. 
\end{enumerate}




\label{algorithms}
\section{Algorithms}\label{sec:algorithms}

\begin{small}
\begin{table*}[t]
\centering
\small
\caption{An overview of negative sampling methods collected from various domains. For acronyms used, ``S" represents Static NS; ``H" refers to Hard NS; ``G" means GAN-based NS; ``A" means Auxiliary-based NS; ``B" represents In-batch NS.}
\label{dataset}
\renewcommand{\arraystretch}{1.25}
\setlength{\tabcolsep}{0.2mm}{
\begin{tabular}{c|c|c|c}  
\hline
Cate & Subcategory & Model Recipe  & Candidates   \\
\hline
\multirow{4}{*}{\tabincell{c}{S}}               &  \multirow{2}{*}{PNS} & Word2Vec~\cite{mikolov2013distributed}$^{(NLP)}$, Deepwalk~\cite{perozzi2014deepwalk}$^{(GRL)}$    & \multirow{4}{*}{Global}  \\
&    &      LINE~\cite{tang2015line}$^{(GRL)}$,Node2vec~\cite{grover2016node2vec}$^{(GRL)}$       &  \\ 
\cline{2-3}
&  \multirow{2}{*}{RNS}   &      BPR~\cite{rendle2012bpr}$^{(RS)}$, LightGCN~\cite{he2020lightgcn}$^{(RS)}$ &  \\
& & TransE~\cite{bordes2013translating}$^{(KGE)}$, DISTMULT~\cite{yang2014embedding}$^{(KGE)}$,RUBER~\cite{tao2018ruber}$^{(NLP)}$,USR~\cite{ghazarian2019better}$^{(NLP)}$   &  \\
\hline
\multirow{5}{*}{\tabincell{c}{H}}                & \multirow{3}{*}{DNS} & FaceNet~\cite{schroff2015facenet}$^{(CV)}$, Max Sampling~\cite{rao2016noise}$^{(NLP)}$, PinSage~\cite{ying2018graph}$^{(RS)}$,~\cite{shrivastava2016training}, ~\cite{bucher2016hard}, ~\cite{harwood2017smart},~\cite{galanopoulos2021hard}    & Global  \\
&  &   SGA~\cite{chen2018improving}$^{(NLP)}$,DNS~\cite{zhang2013optimizing},AOBPR~\cite{rendle2014improving}$^{(RS)}$,BootEA~\cite{sun2018bootstrapping},Dual-AMN~\cite{mao2021boosting}$^{(KGE)}$   & Local   \\
&  & NSCaching~\cite{zhang2019nscaching}$^{(KGE)}$,ESimCSE~\cite{wu2021esimcse}$^{(NLP)}$,MoCoSE~\cite{cao2022exploring}$^{(NLP)}$,MocoRing~\cite{wu2020conditional}$^{(CV)}$  &  Memory-based \\
\cline{2-4}
& \multirow{2}{*}{Mix-NS} &   MoChi~\cite{kalantidis2020hard}$^{(CV)}$ & Cache  \\ 
&  &   MixGCF~\cite{huang2021mixgcf}$^{(RS)}$, MixKG~\cite{che2022mixkg}$^{(KGE)}$ & Global    \\ 
\hline 
\multirow{6}{*}{\tabincell{c}{G}}  & \multirow{3}{*}{Mining}   & IRGAN~\cite{wang2017irgan}$^{(IR)}$,SeqGAN~\cite{yu2017seqgan}$^{(NLP)}$,ACE~\cite{bose2018adversarial}$^{(NLP)}$  &  \multirow{6}{*}{Global} \\
&  &   KBGAN~\cite{cai2017kbgan}$^{(KGE)}$,IGAN~\cite{wang2018incorporating}$^{(KGE)}$, NMRN~\cite{wang2018neural}$^{(RS)}$  &   \\ 
&  &   GraphGAN~\cite{wang2019learning}$^{(GEL)}$ ,ProGAN~\cite{gao2019progan}$^{(GRL)}$   & \\
\cline{2-3}
& \multirow{2}{*}{Generation}  &   CFGAN~\cite{chae2018cfgan}$^{(RS)}$,AdvIR~\cite{park2019adversarial}$^{(IR)}$,HeGAN~\cite{hu2019adversarial}$^{(GRL)}$,SAN~\cite{gupta2021synthesizing}$^{(NLP)}$   &   \\ 
&  &   NDA-GAN~\cite{sinha2021negative},AdCo~\cite{hu2021adco}, CLAE~\cite{ho2020contrastive},NEGCUT~\cite{wang2021instance},DAML~\cite{duan2018deep}$^{(CV)}$  &  \\ 
\hline
\multirow{7}{*}{\tabincell{c}{A}}  &  \multirow{2}{*}{Graph}   & MCNS~\cite{yang2020understanding}$^{(GRL)}$,SANS~\cite{ahrabian2020structure}$^{(KGE)}$  &  Hop  \\
&  &  GNEG~\cite{zhang2018gneg}$^{(NLP)}$,SamWalker~\cite{chen2019samwalker},KGPolicy~\cite{wang2020reinforced}, DSKReG~\cite{wang2021dskreg}, MixGCF~\cite{huang2021mixgcf}$^{(RS)}$  &  Global  \\
\cline{2-4}
& \multirow{2}{*}{Extra} &  SBPR~\cite{zhao2014leveraging}$^{(RS)}$,PRFMC~\cite{manotumruksa2017personalised}$^{(RS)}$,MF-BPR~\cite{loni2016bayesian}$^{(RS)}$, View-aware NS ~\cite{ding2018improved}$^{(RS)}$  & \multirow{2}{*}{Local}    \\
&  &   ReinforcedNS~\cite{ding2019reinforced},RecNS~\cite{yang2022region}  &    \\ 
\cline{2-4}
& \multirow{2}{*}{Cache}  &  Unsupervised Feature Learning~\cite{wu2018unsupervised}$^{(CV)}$,NSCaching~\cite{zhang2019nscaching}$^{(KGE)}$,SRNS~\cite{ding2020simplify}$^{(RS)}$   &  \multirow{2}{*}{Cache}   \\ 
& & MoCo~\cite{he2020momentum}, MoCo-V2~\cite{chen2020improved},MoCoRing~\cite{wu2020conditional}$^{(CV)}$,GCC~\cite{qiu2020gcc}$^{(GRL)}$,ESimCSE~\cite{wu2021esimcse},MoCoSE~\cite{cao2022exploring}$^{(NLP)}$   &  \\
\hline
\multirow{5}{*}{\tabincell{c}{B}}  &  \multirow{2}{*}{Basic}  & N.S.~\cite{chen2017sampling},$S^3$-Rec~\cite{zhou2020s3},SGL~\cite{wu2021self},MHCN~\cite{yu2021self},DHCN~\cite{xia2021self} $^{(RS)}$, SimCLR~\cite{chen2020simple}$^{(CV)}$   &  \multirow{5}{*}{Mini-batch}      \\
&  &   MVGRL~\cite{hassani2020contrastive},GRACE~\cite{zhu2020deep},GraphCL~\cite{you2020graph}$^{(GRL)}$ ,SimCSE~\cite{gao2021simcse},InfoWord~\cite{kong2019mutual}  $^{(NLP)}$  &    \\
\cline{2-3}
& \multirow{1}{*}{Debiased}  &   DCL~\cite{chuang2020debiased}$^{(CV)}$ ,GDCL~\cite{zhao2021graph}$^{(GRL)}$     & \\
\cline{2-3}
& \multirow{2}{*}{Hard}  &   MoCoRing~\cite{wu2020conditional}$^{(CV)}$ ,CuCo~\cite{chu2021cuco},ProGCL~\cite{xia2022progcl}$^{(GRL)}$ ,VaSCL~\cite{zhang2021virtual},\\
& & SNCSE~\cite{wang2022sncse}  $^{(NLP)}$, BatchSampler~\cite{yang2023batchsampler} $^{(NLP,GRL,CV)}$ & \\
    \hline
    \end{tabular}}
\label{tab:ns_algorithms} 
\vspace{-0.4cm}
\end{table*}
\end{small}

In this section, we first summarize five categories of selection methods to form negative sample candidates (i.e., where to sample from?). Next, we summarize a variety of negative sampling algorithms into five categories (i.e., how to sample?).

\subsection{Negative Sample Candidates Construction}\label{sec:candidates}
The process of constructing negative sample candidates is a critical first step in the negative sampling pipeline, determining the pool $\mathcal{C}$ from which negatives are drawn for model training. Here, we answer the first question where should we sample negative examples from? In terms of the composition of the negative sample candidates, we summarize their selection methods into the following five categories.
\begin{itemize}[leftmargin=*]
    \item \textbf{Global Selection} is one of the most common methods for negative sample candidate selection, where the pool of negative samples is composed of all possible negatives from the entire dataset. It ensures diversity in the negative samples but may include less relevant negatives, which could impact the learning efficiency. For example, word2vec~\cite{mikolov2013distributed} utilizes the whole vocabulary as the pool of possible negative samples; LightGCN~\cite{he2020lightgcn} leverages all unobserved items as the pool.

    \item \textbf{Local Selection} focuses on sampling a specific subset of the total available negatives as the pool of negative samples. This method is more selective compared to Global Selection, aiming to construct a pool that is more relevant or challenging for a specific query or anchor. For example, ANCE~\cite{xiong2020approximate} selects top-k negative samples based on the query as the pool of negative samples. Besides, the specific subset can reduce the computational complexity that would be involved in handling the entire set of available negatives. For example, DNS~\cite{zhang2013optimizing} randomly samples a subset as the pool of negatives for probability calculations.

    \item \textbf{Mini-batch Selection} uses other samples in the current mini-batch as the pool without the additional process of choosing. It leverages already-loaded batch data, making it computationally efficient. For example, SimCLR~\cite{chen2020simple} and SimCSE~\cite{gao2021simcse} leverage other samples in the same batch as the pool of negatives.

    \item \textbf{Hop Selection} is a novel selection method for graph-structured data, which selects $k$-hop neighbors as negative sample candidates. This method effectively takes advantage of the graph structure where the information propagation mechanism provides theoretical support. However, matrix operation for obtaining $k$-hop neighbors is impractical for web-scale datasets. Therefore, a path obtained from a random walk or DFS is often considered as a negative sample candidate. For example, RecNS~\cite{yang2022region} selects the intermediate region (i.e., k-hop neighbors) as the pool of negatives for graph-based recommendation; SANS~\cite{ahrabian2020structure} utilizes k-hop neighbors as the pool for knowledge graph embedding.

    \item \textbf{Memory-based Selection} maintains a memory bank or a cache as a pool to store the pool of negative sample candidates. This bank retains a large number of negatives from past iterations or batches, which are used for subsequent training steps. The memory bank is typically updated continuously, with new negatives being added and the oldest ones being removed, ensuring a fresh and diverse set of negative samples. Memory-based selection is particularly advantageous for models that benefit from contrasting a wide array of negatives against positives, such as in contrastive learning scenarios. MoCo~\cite{he2020momentum} uses a queue bank to store the pool of negative samples.
    
\end{itemize}

\subsection{Negative Sampling Algorithms}
Once the negative candidates are constructed, the next step involves negative sampling algorithms, which aim to design a negative distribution (i.e., a specific probability distribution) or a sophisticated negative sampling strategy to sample negative samples from a constructed pool of candidates for model training.

The design of negative sampling algorithms can be simple, such as random sampling, or it can be more sophisticated, taking into account factors like the current state of the model, the difficulty level of the negatives, or their frequency of occurrence in the dataset. Here, we briefly present an overview of negative sampling algorithms. Table ~\ref{tab:ns_algorithms} summarizes existing negative sampling algorithms. The general categorization of negative sampling algorithms and the abbreviated description of each category can be illustrated as follows:
\begin{itemize}[leftmargin=*]
    \item \textbf{Static NS} is a static negative sampling method that keeps a fixed negative sampling distribution throughout the training process. As a basic negative sampling method, static NS is usually utilized to estimate the performance of a newly proposed model optimized with negative sampling. In general, static NS is a simple, fast, and easy-implemented but model-independent method.
    
    \item \textbf{Hard NS} is a model-dependent negative sampling method in which the designed probability of selecting a specific negative sample is related to its relevance or difficulty. Here, hard negatives are close to the decision boundary in the feature space, which makes it difficult for the model to distinguish from positive samples. In summary, Hard NS is a dynamic and mutually promoting method where the applied negative sampling distribution varies with model training.  
    
    \item \textbf{GAN-based NS} is an adversarial negative sampling method that utilizes a generative adversarial network (GAN)~\cite{goodfellow2014generative} to sample negatives. This method employs a generator as a negative sampler to select or generate negatives (fake positives) to confuse the discriminator while a trainable encoder usually acts as the discriminator to distinguish positive and negative samples. GAN-based NS engages in a minimax game, mirroring the adversarial training of GANs to refine the selection of negatives.      
     
    \item \textbf{Auxiliary-based NS} utilizes some auxiliary information to sample negatives, such as extra data types, graph structures, or cache mechanisms. Extra-data-based NS incorporates other types of data into negative sampling, such as view data and exposure data collected from E-commerce platforms. Graph-based NS takes advantage of graph information to sample negatives. Cache-based NS maintains a cache to store important candidate negatives with a fixed frequency update mechanism.  
    
    \item  \textbf{In-batch NS} is a common negative sampling method that can boost training efficiency when a large number of negatives are required, which allows for the sharing of negative samples in a batch. The number of negatives $N$ for each positive pair is related to the batch size $B$. In dense retrieval domain, the number of negatives $N$ can be formulated as $N=B-1$ while can be formulated as $N=2(B-1)$ in contrastive learning. However, this approach must carefully manage the potential for false negatives, which could adversely affect model performance. 

\end{itemize}

Next, we successively review each category of negative sampling algorithms in detail and demonstrate their pros and cons.

\subsubsection{\textbf{Static Negative Sampling}}
Static negative sampling (Static NS) stands as a fundamental approach in the realm of negative sampling, comprising of random negative sampling (RNS) and popularity-based negative sampling (PNS). The general distribution can be represented as: $p_n(x^-) = (\frac{\#x^-}{\sum_{x' \in \mathcal{C}}\#x'})^\beta$ where $\beta$ is a hyperparameter that controls the sampling distribution, $\#x'$ is the frequency of sample $x'$.

\begin{figure}[htbp]
  \centering
  \includegraphics[width=0.44\textwidth]{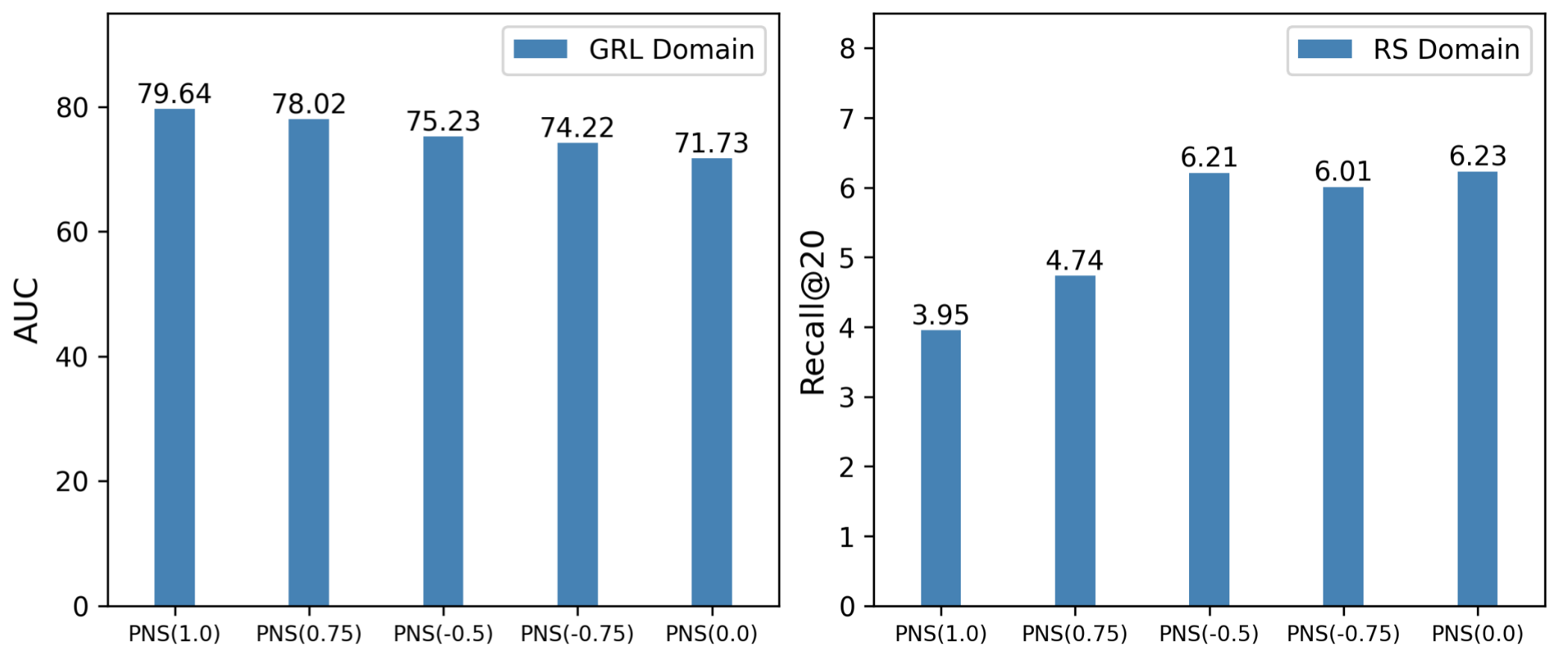}
  \vspace{-0.3cm}
  \caption{Performance comparison with different choices of $\beta$ on GRL and RS domains, respectively.}
  \label{fig: sns_perf}
\end{figure}

\begin{itemize}[leftmargin=*]
    \item \textbf{Random Negative Sampling (RNS).} As the most prevalent negative sampling method, RNS is widely used as the default sampling method to evaluate the effectiveness of the proposed model in optimization with negative sampling. RNS designs uniform sampling weights for each negative sample when deciding which ones to use during training. Uniform weights mean that every negative sample has an equal chance of being selected, implying no preference among the negatives. However, its indiscriminate sampling fails to prioritize informative negatives, potentially diluting the training signal with less relevant samples.

    \item \textbf{Popularity-based Negative Sampling (PNS).} As proposed in previous work~\cite{pan2008one}, user-oriented and item-oriented sampling methods substitute uniform sampling by counting the number of interactions of users and items, respectively. It exploits data properties to sample negatives in early recommendation efforts. Word2vec~\cite{mikolov2013distributed} empirically sets the negative sampling distribution as the $3/4$ power of word frequency for faster training and better word embedding learning. Most later studies on network embedding~\cite{perozzi2014deepwalk, tang2015line, grover2016node2vec} or graph representation learning~\cite{hamilton2017inductive} also keep this setting for negative sampling. However, the most appropriate choice of $\beta$ varies with different datasets and fields (See Figure~\ref{fig: sns_perf}). Results in ~\cite{caselles2018word2vec, zhang2018gneg} demonstrate that performance shows a strong dependency on the choice of $\beta$.
    
\end{itemize}

\noindent{\textbf{Pros and Cons.}} 
Static NS is a simple and easy-implemented negative sampling method, making it an accessible option for a wide range of applications. Compared with other heuristic sampling methods, Static NS has the characteristics of being fast, stable, and model-independent. A significant limitation of these methods is that the negative sampling distribution remains static throughout the training process, failing to adjust the selection of negative samples to suit each positive pair and resulting in sampling low-quality negatives. In general, performance increases as the number of negatives increases. However, too many negatives lead to an increase in GPU memory and training time. Therefore, the number of negative samples should be further investigated to accomplish the trade-off between performance and time consumption.

\subsubsection{\textbf{Hard Negative Sampling}} 
Hard Negative Sampling (Hard NS) addresses the limitations of static NS by specifically targeting hard negatives. A concept of hard negative examples was first proposed in the bootstrapping method~\cite{rowley1995human, sung1998example}, which treated incorrectly classified examples as hard examples. Over time, this idea evolved to not just include misclassified examples, but also those samples that are close to the decision boundary in the feature space, irrespective of whether they have been previously misclassified or not. Thus, hard negatives are close to the decision boundary in the embedding space, which is difficult for a model to distinguish from positives.

Hard negative sampling dynamically selects hard negatives based on proximity in the feature space or based on the confidence level of the model’s predictions. Based on the source of negatives, hard negative sampling can be divided into dynamic negative sampling (DNS) and mixture-based DNS. DNS dynamically mines hard negatives from the raw data while mixture-based DNS directly synthesizes negative embedding from the embedding space.

\begin{itemize}[leftmargin=*]
    \item \textbf{Dynamic Negative Sampling (DNS).} In DNS, negative samples are not predetermined but are selected based on the current state of the model or specific criteria that evolve as the training progresses. In light of the criteria, DNS can be divided into three groups, including anchor-based sampling, positive-based sampling, and anchor-positive sampling. 
\end{itemize}

\begin{enumerate}[label=(\arabic*)]
    \item \noindent \textbf{Anchor-based DNS.} This group focuses on selecting negatives based on their relationship or similarity to a given anchor. The anchor acts as a reference, and negatives are chosen based on how they differ from this specific anchor in the feature space. The general distribution can be represented as $p_n(x^-) = \frac{S(f(x), f(x^-))}{\sum_{x'\in \mathcal{C}}S(f(x), f(x'))}$. Rendle et al.~\cite{rendle2014improving} developed an adaptive and efficient item sampler to select informative negative items with a small predicted rank, where the rank is designed as context-dependent and obtained by the score model among all items. Chen et al.~\cite{chen2018improving} proposed an alternative sampler to replace word-popularity-based sampling in the skip-gram model, which prefers to select negative samples with higher inner product scores. Wu et.al~\cite{wu2017sampling} designed a distance-weighted sampling method to sample informative negatives where the distance weights between the anchor and negatives are clipped to correct the bias. In addition to inner product and distance, PageRank score can also be used to measure similarity in PinSage~\cite{ying2018graph}.

    \item \noindent \textbf{Positive-based DNS.} Although anchor-based DNS offers better convergence and performance, it cannot take the impact of positive samples into consideration. It may seem intuitive that sampled negatives that are closer to positive samples will provide a sufficient risk of discrimination in order to distinguish positive from negative samples. The general distribution can be represented as $p_n(x^-) = \frac{S(f(x^+), f(x^-))}{\sum_{x'\in \mathcal{C} } S(f(x^+), f(x'))}$ . Max Sampling~\cite{rao2016noise} selected the most difficult negative answer by maximizing the similarities between the positive one and all negatives. Tran et al.~\cite{tran2019improving} proposed a 2-stage sampling method, which first sampled a small subset of negative candidates and then selected informative negatives based on the similarity between a positive item and the candidates. Sun et al.~\cite{sun2018bootstrapping} developed a $\epsilon$-Truncated uniform negative sampling method to mine hard negative in KGE by using $s$-nearest neighbors of positive entities as candidates.

    \item \noindent \textbf{Anchor-Positive DNS.} This method involves choosing negatives based on their joint relationship with both the anchor and the positive samples, which is valuable in complex learning scenarios like hinge loss frameworks. 
    The general distribution can be represented as $p_n(x^-)=\frac{S(f(x), f(x^+), f(x^-))}{\sum_{x'\in \mathcal{C} } S(f(x), f(x^+), f(x'))}$. 
    In computer vision, several works~\cite{simo2015discriminative, loshchilov2015online, wang2015unsupervised, shrivastava2016training} selected hard negative examples based on the triple loss, where negatives are close to the anchor and incur a high value of the loss. For WARP loss (Weighted Approximate-Rank Pairwise loss), a myriad of efforts~\cite{weston2011wsabie, zhao2015improving} adopted uniform negative sampling with rejection to mine hard negatives, which uniformly sampled a negative example until satisfied the constraint of $1-S(f(x), f(x^+))+S(f(x), f(x^-))>0$. Guo et al.~\cite{guo2018approximating} also designed a dynamic sampler in word embedding to select informative negative words with higher ranking scores than positive ones. Faghri et al.~\cite{faghri2017vse++} combined anchor-based DNS and positive-based DNS to sample hard negative examples within each mini-batch in the image-caption retrieval domain. Guo et al.~\cite{guo2018vse} selected high-ranked negative labels by a rank-invariant transformation for fast sampling. To improve the robustness, Several works~\cite{schroff2015facenet,li2019sampling} proposed a semi-hard negative sampling method to select ``semi-hard'' negatives that were farther from the anchor-positive pair with a distance margin. 

\end{enumerate}

\begin{itemize}[leftmargin=*]
    \item \noindent \textbf{Mixture-based DNS.} Inspired by Mixup~\cite{zhang2017mixup}, the idea of synthesizing hard negative in embedding space has become one of the popular negative sampling methods. Unlike methods that select samples existing in the dataset as negatives, Mixture-based DNS works at the level of embeddings, creating synthetic negative samples by strategically combining embeddings among several selected negatives. The synthetic negative samples can be represented as $\textbf{x}^{-}=\text{Combine}(\{\textbf{x}_i\})$, where $\textbf{x}_i$ is a vector embedding of a selected sample $x_i$ and $x_i \sim p_n(x^-)$. The function $\text{Combine}(\cdot)$ is utilized to combine the selected negatives for synthetic negative samples.
    In contrastive learning, MoChi~\cite{kalantidis2020hard} designed a hard negative mixing method to synthesize harder negatives by mixing the query and the hardest negatives in embedding space where the hardest ones are generated by anchor-based DNS. MixGCF~\cite{huang2021mixgcf} utilized the information propagation mechanism to design a hop-mixing strategy to synthesize negatives from multiple hops in a graph and adopted positive mixing to enhance negatives.  MixKG~\cite{che2022mixkg} mixed candidate negative samples via the convex combination operation where candidate ones were filtered by two criteria: score-function based sampling and entity similarity correcting. 
\end{itemize}

\begin{figure}[t]
  \centering
  \includegraphics[width=0.48\textwidth]{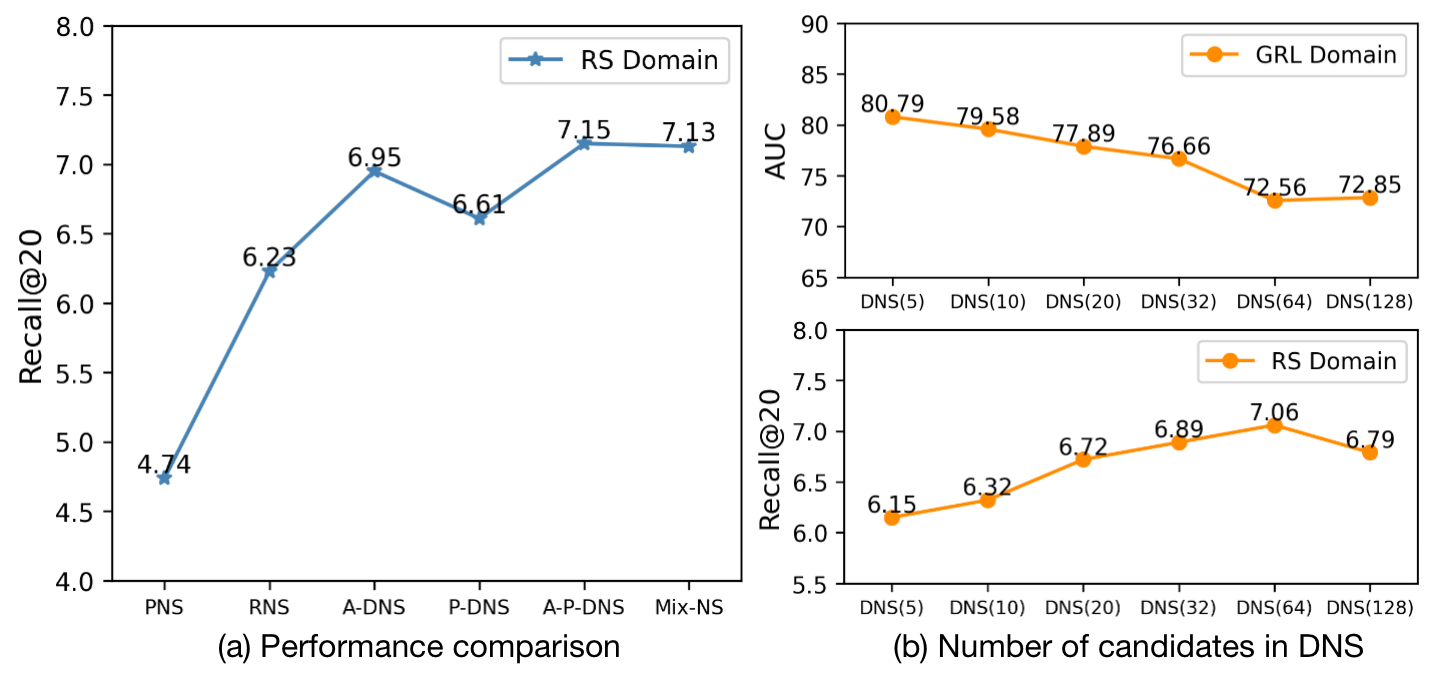}
  \vspace{-0.3cm}
  \caption{(a) Performance comparison between Static NS and Hard NS; (b) Impact of the number of negative candidates in DNS.}
  \label{fig: hard_result}
  \vspace{-0.4cm}
\end{figure}

\noindent \textbf{Pros and Cons.} Hard NS is a model-dependent negative sampling technique that dynamically samples hard negative examples, which accelerates convergence and gives a clear gradient update direction for model training. Hard NS achieves significant performance improvements compared with static NS (e.g. PNS and RNS) (See Figure~\ref{fig: hard_result} (a)). However, such methods have an obvious disadvantage, namely that require more time to obtain distribution by the current model. To improve the efficiency, hard negative sampling usually samples a subset or takes mini-batch examples as negative candidates. Thus, the effectiveness of Hard NS relies on the number of candidates, which varies largely on different datasets (See Figure~\ref{fig: hard_result} (b)). More importantly, false negatives have gradually attracted more attention since they are closer to positive samples in embedding space and are difficult to distinguish from true negative samples. Many works~\cite{chuang2020debiased, ding2020simplify, chen2021incremental, yang2022trading} focus on alleviating the false negative issue and boosting the robustness of the sampling process. Overall, Hard NS is an effective negative sampling method to speed up convergence and boost performance.

\subsubsection{\textbf{GAN-based Negative Sampling}}
GAN-based negative sampling methods utilize generative adversarial networks (GANs) to mine or generate adversarial negatives, which are widely adopted in multiple domains. For GAN-based NS, the generator serves as a negative sampler to mine or generate informative negatives (false positives) to confuse the discriminator. The trainable models serve as a discriminator to distinguish the true positives and false positives (negative). GAN-based NS methods can be roughly divided into two subcategories: GAN-based negative mining and GAN-based negative generation.

\begin{itemize}[leftmargin=*]
    \item \textbf{GAN-based Negative Mining.} GAN-based negative mining utilizes GAN to adaptively mine informative negatives from the dataset. This method relies on discrete indexes of raw data, with the generator identifying potential negatives to challenge the discriminator. The general distribution can be represented as $p_n^G(x^-)=\frac{S(G(x), G(x^-))}{\sum_{x' \in \mathcal{C}} S(G(x), G(x'))}$ where $G$ denotes a generator that generates negatives. SeqGAN~\cite{yu2017seqgan} is the first work to utilize the generator to select discrete negatives and directly apply policy gradient to achieve gradient passing from the discriminator to the generator. IRGAN~\cite{wang2017irgan} utilized the generative information retrieval model to select the discrete index of samples as negatives, which uses REINFORCE~\cite{williams1992simple} for model optimization. The abovementioned methods provide a new idea for negative sampling that designs an adversarial negative sampler and applies a policy gradient-based reinforcement learning (RL) method for model optimization. Such negative sampling methods are widely developed in other fields. For example, GraphGAN~\cite{wang2019learning} employed a GAN-based framework for graph representation learning, and NMRN~\cite{wang2018neural} proposed an adaptive negative sampler based on GAN in a streaming recommendation, and ACE~\cite{bose2018adversarial} adopted GAN to find harder negatives in word embedding. In knowledge graph embedding, KBGAN~\cite{cai2017kbgan} utilized one of the existing knowledge graph embedding models as the generator to sample high-quality negatives while Wang et al.~\cite{wang2018incorporating} employed a two-layer fully-connected neural network as the generator for negative sampling.

    \item \textbf{GAN-based Negative Generation.} Instead of mining negatives from raw data, GAN-based negative generation aims to generate synthetic negative embedding directly from the embedding space for model optimization. This method generates hard negatives by synthesizing new samples that are specifically designed to challenge the discriminator. The general distribution can be represented as  
    $p_n(\textbf{x}^-)=\frac{S(\textbf{x}, \textbf{x}^{-})}{\sum_{x' \in \mathcal{C}} S(\textbf{x}, \textbf{x'})}$, where $\textbf{x}$, $\textbf{x}^{-}$, and $\textbf{x'}$ are generated vectors by the generator $G$. CFGAN~\cite{chae2018cfgan} utilized the generator to generate continuous negative embeddings composed of real-valued elements in collaborative filtering without using the RL method for model optimization. DAML~\cite{duan2018deep} employed adversarial learning to synthesize hard negatives from inputs (anchor, positive, and negative) by the generator for metric learning. Different from IRGAN comprising of two models, AdvIR~\cite{park2019adversarial} only used a single model to combine adversarial sampling and adversarial training to generate hard negative examples by adding adversarial perturbation vectors on negative example vectors. Similar to AdvIR, CLAE~\cite{ho2020contrastive} designed an adversarial training algorithm for self-supervised learning that leveraged adversarial examples to generate harder negatives. HeGAN~\cite{hu2019adversarial} extended GAN-based NS into Heterogeneous Information Networks (HIN), which generates ``latent" negative nodes from a continuous distribution rather than true existing nodes. In dialogue systems, Gupta et al.~\cite{gupta2021synthesizing} synthesized adversarial negative responses. AdCo~\cite{hu2021adco} applied GAN-based NS to contrastive learning to generate challenging negative examples. Sinha et al.~\cite{sinha2021negative} proposed negative data augmentation (NDA) and integrated it into GAN where NDA acted as the generator to synthesize hard negatives for the discriminator. Chen et al.~\cite{chen2021novelty} utilized negative data augmentation (NDA) for novelty detection.

\end{itemize}

\begin{figure}[h]
  \centering
  \includegraphics[width=0.47\textwidth]{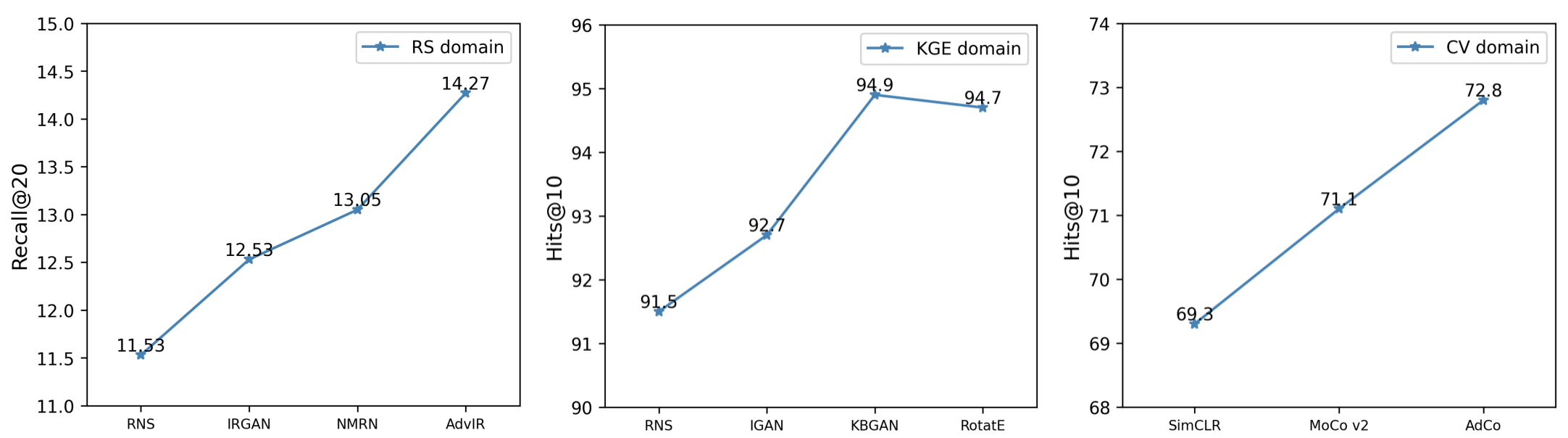}
  \vspace{-0.3cm}
  \caption{Performance comparison between different GAN-based NS and RNS on recommendation system (left), knowledge graph embedding (middle), and computer vision (right).}
  \label{fig: ns_gan}
  \vspace{-0.4cm}
\end{figure}

\noindent \textbf{Pros and Cons.} 
GAN-based NS leverages adversarial learning to adaptively mine or generate negative samples, which provides more informative negatives and achieves faster convergence and better performance. This method adopts an adaptive negative sampler as a generator for sufficiently exploiting training samples or embedding space to seek more informative negatives. 
As shown in Figure~\ref{fig: ns_gan}, compared to RNS, GAN-based NS achieves better performance on different domains since it mines more informative negatives for model training. However, GAN-based NS has an obvious shortcoming of unstable training. A common method to alleviate this issue is to pretrain the discriminator and the generator, which violates the original purpose of using negative sampling to speed up training. Thus, a stable optimization for GAN-based NS is a core and challenging problem, which leaves much room to investigate. Besides, GAN-based NS usually requires an additional network as the generator, which increases the training time and computational load. In general, GAN-based NS has a promising prospect for discovering hard negative samples but it is time-consuming, limiting its application to large-scale datasets.

\subsubsection{\textbf{Auxiliary-based Negative Sampling.}} Auxiliary-based negative sampling methods rely on auxiliary information to sample negatives, comprising extra-data-based NS, graph-based NS, and cache-based NS. 
Each subcategory of Auxiliary-based NS possesses its unique characteristics, which will be demonstrated in detail. 

\begin{itemize}[leftmargin=*]
    \item \textbf{Extra-data-based NS.} This method utilizes external datasets or additional information beyond primary training set, enriching negative samples with a broader context or more examples and allowing for the selection of negatives that are more informative or challenging. For example, within the realm of recommendation systems, leveraging additional sources of information, such as social links, view data, and exposure data, can significantly enrich the negative sampling process. These auxiliary data types are particularly valuable as they offer insights into user preferences and behaviors that are not captured by traditional interaction data alone.
    SBPR~\cite{zhao2014leveraging} incorporated social links into negative sampling where social feedback served as negatives for positive items but was regarded as positives for unobserved negative items. Instead of utilizing a single additional information for negative sampling, PRFMC~\cite{manotumruksa2017personalised} leveraged both social links and geography to enhance negative sampling. MF-BPR~\cite{loni2016bayesian} utilized multi-feedback data to compose negative items, which proposed a non-uniform negative sampler that quantified the impact of different types of feedback. Mix-Exp-NS~\cite{ding2018improved} integrated view data into negative sampling, which designed a sampling weight to sample negatives from view and unobserved items. RNS-AS~\cite{ding2019reinforced} combined exposure data and adversarial training into negative sampling to select high-quality real negatives. RecNS~\cite{yang2022region} mixed negatives from positive-assisted and exposure-enhanced negative sampling methods to select hard and real negatives.

    \item \textbf{Graph-based NS.} This method capitalizes on the rich structural information inherent in graph data to enhance negative sampling. This approach is particularly pertinent in scenarios where data can be naturally represented as graphs, such as social networks, citation networks, or any domain where relationships between entities can be constructed as a graph. GNEG~\cite{zhang2018gneg} leveraged random walk on the constructed word co-occurrence network to obtain negative sampling distribution. SamWalker~\cite{chen2019samwalker} conducted the personalized random walk with a walking probability on a social network to obtain the distribution for sampling informative negative examples. SamWalker++~\cite{wang2021samwalker++} designed a pseudo-social network to substitute the additional social network for random walk-based negative sampling. MCNS~\cite{yang2020understanding} leveraged self-contrast approximation to closely mimic the distribution of positive distribution and utilized Metropolis-Hastings to accelerate negative sampling based on graph structure, which adopted DFS to traverse the graph for generating Markov chain. KGPolicy~\cite{wang2020reinforced} integrated item knowledge graph into recommendation for negative sampling, which adopted reinforcement learning framework to select informative negative samples. MixGCF~\cite{huang2021mixgcf} utilized graph structure to sample hard negative items from multiple layers and mixed these with the positive item to obtain the synthetic negatives.

    \item \textbf{Cache-based NS.} This method employs a cache or  memory bank as a dynamic repository for negative samples. In this way, the model can quickly and efficiently access these samples during training. This approach significantly reduces the computational overhead associated with sampling new negatives for each training iteration. Wu et al.~\cite{wu2018unsupervised} maintained a memory bank for storing negative samples. NSCaching~\cite{zhang2019nscaching} used a cache to maintain rare hard negative triplets and uniformly selected negatives from the cache for knowledge graph embedding learning. SRNS~\cite{ding2020simplify} designed a memory-based negative sampler to store high-variance negative candidates, which can effectively alleviate the false negative issue in recommendation. In unsupervised visual representation learning, MoCo~\cite{he2020momentum} utilized a queue to store mini-batches data as negatives for the next iteration. MoCHi~\cite{kalantidis2020hard} and MoCoRing~\cite{wu2020conditional} utilize hard negative sampling methods to enhance Cache-based NS. 
    GCC~\cite{qiu2020gcc} maintained a queue to store negative instances for unsupervised graph representation learning. Cert~\cite{fang2020cert} utilized a queue to store negative examples for language understanding. ESimCSE~\cite{wu2021esimcse} and MoCoSE~\cite{cao2022exploring} leveraged a negative sample queue to further improve sentence embedding learning. Cache-based NS exploits more negatives during the training but requires more time to update the cache. Thus, a fast cache update mechanism should be explored to further improve the efficiency of cache-based NS.

\end{itemize}

\noindent \textbf{Pros and Cons.} Auxiliary-based NS enriches the process of selecting negative samples by incorporating a variety of auxiliary information, enhancing the quality of negative samples. However, it's important to consider the trade-offs in terms of efficiency and domain-specific applicability. For example, the need to collect extra data can be a limitation, as it may not be feasible or practical in all domains. Additionally, the increased data volume can lead to higher data loading overheads, impacting computational efficiency. Graph-based NS depends on graph algorithms for traversing and mining negatives, which may increase the time consumption.
Cache-based NS eliminates the constraint of GPU memory size for In-batch NS, but the update mechanism of the cache is very time-consuming. Thus, a trade-off between effectiveness and efficiency should be considered in negative sampling methods.

\subsubsection{\textbf{In-batch Negative Sampling}}
In-batch negative sampling method is a distinct approach within the realm of negative sampling techniques, primarily relying on the composition of mini-batch data during training. Unlike other negative sampling methods that explicitly sample negatives for each positive pair, In-batch NS leverages the inherent structure of mini-batches to implicitly sample negatives. This method is particularly notable for its efficiency and simplicity. In-batch NS can be divided into three subcategories, including basic, debiased, and hard.

\begin{itemize}[leftmargin=*]
    \item \textbf{Basic In-batch NS.} This method is first proposed for neural network-based collaborative filtering~\cite{chen2017sampling} to utilize the non-linked examples within the same mini-batch as negatives. It has been effectively adopted in various domains, from image embedding to dual-encoder models in NLP, where it significantly increases the number of negative examples with minimal overhead. Ye et al.~\cite{ye2019unsupervised} leveraged In-batch NS to optimize the negative log-likelihood objective for unsupervised image embedding learning. In a dual-encoder model, In-batch NS has been used as a significant and effective trick for model training where $(B-1)$ examples in the same mini-batch are regarded as negatives~\cite{yi2019sampling, karpukhin2020dense, qu2020rocketqa}, which efficiently boosts the number of negative training examples. In contrastive learning, SimCLR~\cite{chen2020simple} leveraged In-batch NS for visual representation learning where $2(B-1)$ examples generated by data augmentation are treated as negatives. 
    Later, such a simple framework is widely used in various domains for unsupervised representation learning. SimCSE~\cite{gao2021simcse} follows SimCLR framework for sentence embedding learning. Several prior works~\cite{hassani2020contrastive, zhu2020deep, you2020graph, zhu2021graph} on graph contrastive learning also adapt SimCLR framework to design augmentation transformations based on graph structure. SGL~\cite{wu2021self} incorporated self-supervised graph learning into recommendation and designed two types of In-batch NS for the auxiliary task.

    \item \textbf{Debiased In-batch NS.} A primary bias introduced by In-batch NS is false negatives, which can significantly affect the performance of the model. In basic In-batch NS, samples within a mini-batch that are not explicitly labeled as positive are automatically treated as negatives. This assumption can lead to the inclusion of false negatives -- actual positive samples mistakenly treated as negatives due to the absence of explicit positive labeling in the batch. DCL~\cite{chuang2020debiased} proposed a debiased contrastive objective to alleviate sampling bias, which corrected the weights of negatives with positive examples in InfoNCE. DCLR~\cite{zhou2022debiased} designed a debiased objective function that adjusted the weights of negatives using the complementary model. Huynh et al.~\cite{huynh2022boosting} mitigated the effect of false negatives by two approaches: false negative elimination and attraction, which identified false negatives and removed them from the original negative candidates and added them into positive pairs set.

    \item \textbf{Hard In-batch NS.} Basic In-batch NS typically treats all samples within a mini-batch as negatives, assuming these samples are randomly sampled from the training data. However, this approach ignores the importance of the difficulty level of negative samples. Hard In-batch NS method seeks to identify and utilize hard negatives within the batch.   
    HCL~\cite{robinson2020contrastive} developed a hard negative sampling method for contrastive learning, which reweights the weights of negatives in the objective. Xiong et al.~\cite{xiong2020approximate} globally selected hard negative samples by an asynchronously updated ANN index, in which negatives were generated by the dense retrieval (DR) model. Zhan et al.~\cite{zhan2021optimizing} adopted dynamic hard negatives to boost the training process and the ranking performance of the DR model. Yang et al.~\cite{yang2023batchsampler} proposed BatchSampler to globally sample hard negatives for contrastive learning.

\end{itemize}

\begin{figure}[h]
  \centering
  \vspace{-0.3cm}
  \includegraphics[width=0.45\textwidth]{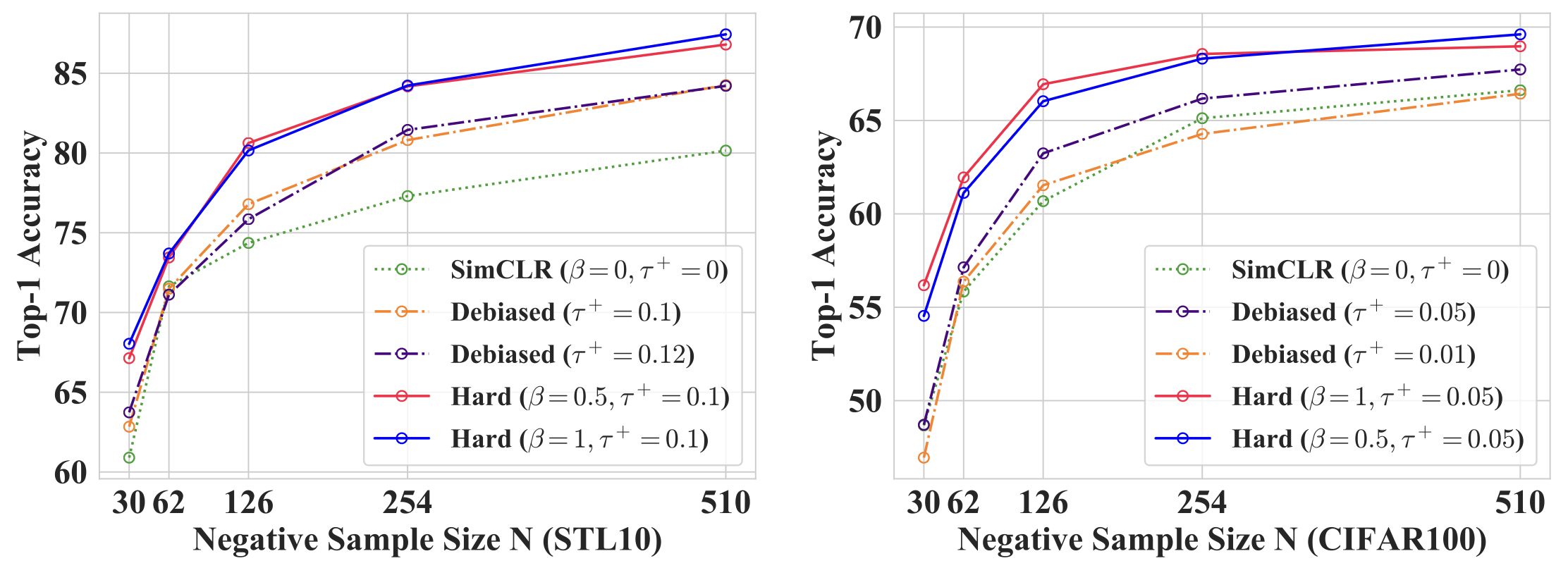}
  \caption{Classification performance comparison with different In-batch NS on top-1 accuracy. Taken from ~\cite{robinson2020contrastive}}
  \label{fig: in-batch-perf}
  \vspace{-0.3cm}
\end{figure}

\noindent \textbf{Pros and Cons.} In-batch NS stands out as an effective and efficient approach, particularly due to its ability to reuse samples from the current mini-batch without necessitating additional sampling operations. This method benefits from larger batch sizes, allowing for a broader range of negative samples to be included in training, which is crucial for effective learning in contrastive learning frameworks. However, this means that the effectiveness of In-batch NS depends on the batch size. As indicated in Figure~\ref{fig: in-batch-perf}, the downstream performance exhibits a noticeable variation in relation to the batch size used during training. There is a clear trend indicating that larger batch sizes can improve downstream performance. In fact, In-batch NS is equivalent to random sampling since the data in the mini-batch are generated randomly. Compared to the basic In-batch NS leveraged by SimCLR, Debiased In-batch NS and Hard In-batch NS achieve better performance (See Figure~\ref{fig: in-batch-perf}). Thus, integrating a hard negative mining strategy into In-batch NS is a huge challenge, which leaves us a lot of room for exploration. Furthermore, since In-batch NS is usually applied in contrastive learning for unsupervised representation learning, the false negative issue is an inevitable problem that plays a significant effect on the downstream performance.

\label{application}
\section{Applications}\label{sec:application}
Negative sampling is an essential technique, which has been widely used in various domains(e.g. recommendation, graph representation learning, knowledge graph embedding, natural language processing, and computer vision).

\subsection{Negative Sampling in RS}
Recommender Systems (RS) have emerged as an indispensable tool for information filtering across a variety of online platforms, including e-commerce sites, advertising platforms, and entertainment services. The core challenge in developing effective RS lies in accurately modeling user preferences, a task primarily reliant on interpreting historical interaction data. Such data often records positive user interactions -- like purchases, likes, or views -- but seldom includes explicit negative feedback, making it difficult to discern user dislikes or indifference. To tackle this, negative sampling has been widely applied in previous works~\cite{pan2008one, rendle2012bpr,he2020lightgcn,huang2021mixgcf}. 
\begin{figure}[htbp]
  \centering
  \vspace{-0.4cm}
  \includegraphics[width=0.44\textwidth]{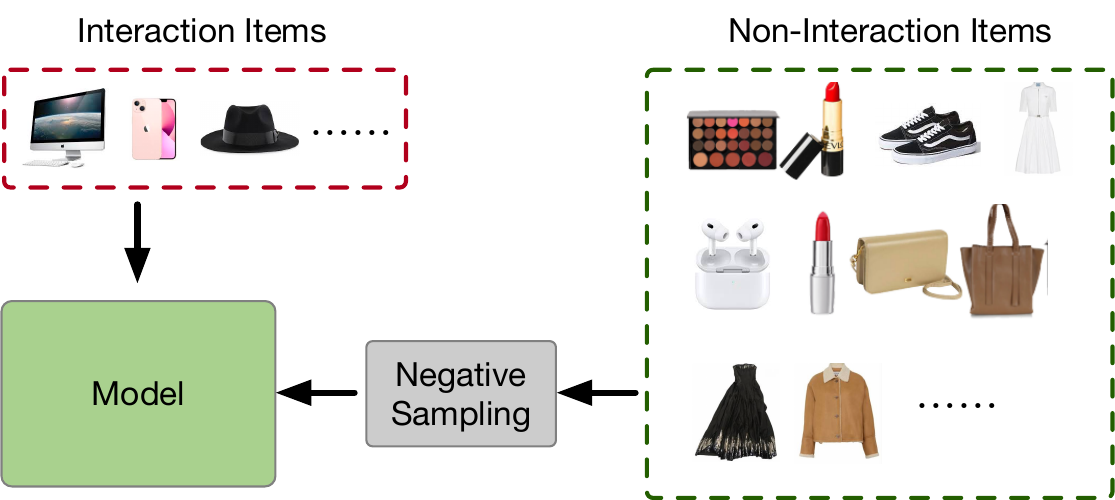}
  \vspace{-0.3cm}
  \caption{The process of negative sampling in a recommender system where negative samples are sampled from non-interaction items.}  
  \label{fig: ns_in_rs}
  \vspace{-0.2cm}
\end{figure}

As shown in Figure~\ref{fig: ns_in_rs}, negative sampling aims to sample a small portion of items from the non-interaction items. The most popular negative sampling method is random negative sampling (RNS)~\cite{pan2008one, rendle2012bpr}, which serves as a basic negative sampling method to evaluate the performance of the proposed model. However, RNS assigns equal weights to unobserved items, which usually draws uninformative negative items for model training. To improve the quality of negative samples, many previous works mainly focus on popularity-based negative sampling (PNS)~\cite{barkan2016item2vec, yu2017selection, lian2020personalized}, hard negative sampling (hard NS)~\cite{zhang2013optimizing, rendle2014improving}, and GAN-based negative sampling (GAN-based NS)~\cite{jin2020sampling, wang2018neural}. The high-quality negative items contribute more to the gradient of the loss function, increasing its magnitude and accelerating convergence. PNS selects negative items based on item popularity, which depends on the data distribution. Intuitively, popular items that are non-interacted with a user are more likely to be negative. However, results in~\cite{caselles2018word2vec} reveal that the smoothing parameter for negative sampling has a significant effect on recommendation performances.
Hard NS methods draw negative items based on prediction scores between the users and candidate negative items. Such methods aim to sample hard negative items with higher prediction scores. Additionally, GAN-based NS methods use the generator to obtain the sampling distribution for adaptively sampling informative negative items.
However, the abovementioned methods suffer from the false negative issue where more hard (informative) negative items are more likely to be positive items. Synthesizing hard negative items has become an effective way to address this problem, which synthesizes or generates negatives in embedding space rather than sampling an existing item from raw data. 

Graph Neural Networks (GNNs) offer a fresh perspective on RS by modeling user-item interactions as bipartite graphs~\cite{ying2018graph, berg2017graph, wang2019neural, he2020lightgcn} or session graphs~\cite{wu2019session}, enabling the leveraging of graph structures for more insightful recommendations. In light of this, graph learning methods provide a novel perspective for recommender systems, which incurs a myriad of GNN-based recommendation models in recent years. Similar to the traditional recommendation, negative sampling is also applied for model training in GNN-based recommendation models. Such sampling methods can incorporate graph information to sample negative items. For example, Pinsage~\cite{ying2018graph} sampled hard negative items based on personalized PageRank scores for the Pinterest recommendation. MCNS~\cite{yang2020understanding} performed DFS on the graph to generate a Markov chain for negative sampling. KGPolicy~\cite{wang2020reinforced} integrated a knowledge graph into negative sampling for seeking high-quality negative items. MixGCF~\cite{huang2021mixgcf} proposed hop mixing which mixed negative items sampled from different hops on the graph. To sum up, negative sampling in GNN-based recommendation can leverage graph information to design better negative sampling strategies but the efficiency issue should also be considered.     

In summary, negative sampling has been proven to be an effective way to improve recommendation performance. As it applies to online services, recommendation systems must seriously consider the efficiency of negative sampling. In-batch NS is an efficient way that can be applied to large-scale recommendation. Thus, a challenging direction is how to reduce the sampling bias for In-batch NS.

\subsection{Negative Sampling in GRL}
Graph representation learning (GRL) has received a myriad of attention in recent years due to its effectiveness in handling graph-structured data. This data structure is prevalent across a wide range of applications, including  social networks, biological networks, academic networks, and many other domain-specific networks. The essence of GRL lies in its ability to encode the complex structure and features of graphs into low-dimensional embeddings, facilitating their use in downstream tasks such as node classification, link prediction, graph classification, and clustering.

\begin{figure}[t]
  \centering
  \includegraphics[width=0.4\textwidth]{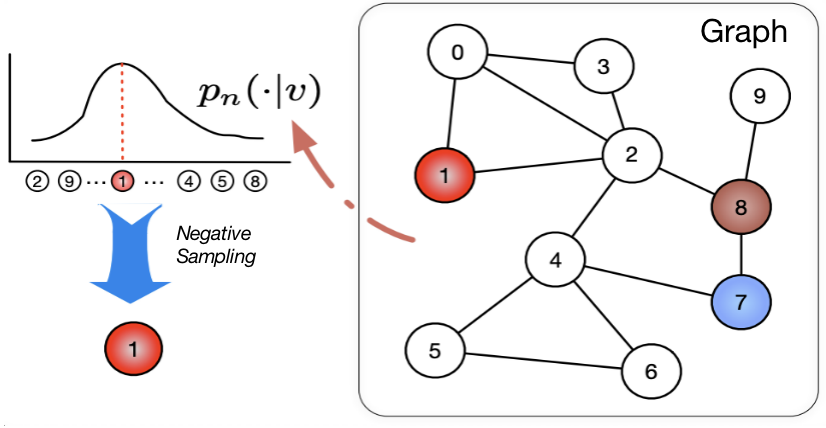}
  \vspace{-0.3cm}
  \caption{The process of negative Sampling in GRL. Taken from ~\cite{yang2020understanding}.}  
  \label{fig: ns_in_graph}
  \vspace{-0.4cm}
\end{figure}

Given the computational challenge of processing all nodes within large-scale graphs, negative sampling emerges as a crucial strategy to enhance efficiency by selecting a subset of nodes as negatives based on the noise distribution learned from the graph (See Figure~\ref{fig: ns_in_graph}). Early implementations in GRL, such as DeepWalk~\cite{perozzi2014deepwalk} and Node2vec~\cite{grover2016node2vec}, followed the negative sampling setting in word2vec, which selected negative nodes according to the empirical unigram distribution proportional to the 3/4 power. However, this predefined sampling distribution can not dynamically sample negative nodes according to the training process. Gao et al.~\cite{gao2018self} proposed a self-paced negative sampling strategy to gradually sample the informative negative nodes for model optimization. Inspired by IRGAN, Gao et al.~\cite{gao2018self} incorporated generative adversarial network (GAN) into self-paced negative sampling to form an extension version of adversarial self-paced negative sampling. Robust-NS~\cite{armandpour2019robust} argued that popularity-based negative sampling failed to accurately estimate the objective of skip-gram due to the popular neighbor problem, which proposed a distance-based negative sampler to draw negative nodes from candidate nodes without neighbors. As a most popular GNNs-based method, GraphSage~\cite{hamilton2017inductive} utilized negative sampling to optimize a graph-based loss function, which also kept the setting of word2vec. GraphGAN~\cite{wang2019learning} integrated GAN into graph representation learning where negative samples were sampled by the generator. Later, MCNS~\cite{yang2020understanding} systematically analyzed the role of negative sampling in graph representation learning and proposed Markov chain Monte Carlo negative sampling.

Recently, contrastive learning has attracted a surge of interest for unsupervised visual representation learning~\cite{wu2018unsupervised, he2020momentum, chen2020simple}. In terms of the great success of contrastive learning in computer vision, numerous works extended it into graph learning~\cite{qiu2020gcc,zhao2021graph}.
After that, negative sampling in graph contrastive learning has also achieved tremendous success. 
Zhao et al.~\cite{zhao2021graph} proposed a graph-debiased contrastive learning framework to alleviate the false negative issue, which utilized clustering to obtain pseudo labels and conducted random sampling to sample negatives from the different clusters. Such a method suffers from expensive computational consumption. CuCo~\cite{chu2021cuco} proposed a curriculum contrastive learning framework that gradually samples negatives from easy to hard, in which hard negatives are determined by the score function. Zhu et al.~\cite{zhu2022structure} utilized heterogeneous graph structure to sample hard negatives with the largest similarities and synthesized more negatives by a mixing operation.
ProGCL~\cite{xia2022progcl} utilized a beta mixture model (BMM) to mine true and hard negative examples for graph contrastive learning. 
Xiong et al.~\cite{xiong2020approximate} proposed approximate nearest neighbor negative contrastive learning (ANCE) for the dense retrieval (DR) model, which globally sampled hard negatives with top retrieved scores from the current DR model.

In summary, negative sampling in graph representation learning has achieved some remarkable progress but still leaves a lot of room for improvement. For example, how to effectively use graph structure for negative sampling; how to incorporate GNNs propagation mechanism into negative sampling? How to design an effective negative sampling method for graph contrastive learning?

\subsection{Negative Sampling in KGE}
Knowledge graph embedding (KGE) provides a new way to represent complex human knowledge in a structured form, encapsulating entities, relationships, and semantic descriptions within a multi-relational graph framework. Each relation in a knowledge graph can be represented as a triple of $<head\ entity, relation, tail\ entity>$. The key idea of knowledge graph embedding is to embed entities and relations in a KG into a continuous embedding space. The learned embeddings pave the way for many downstream applications, including knowledge graph completion, relation extraction, entity discovery, question answering, and recommender systems.

\begin{figure}[t]
  \centering
  \includegraphics[width=0.48\textwidth]{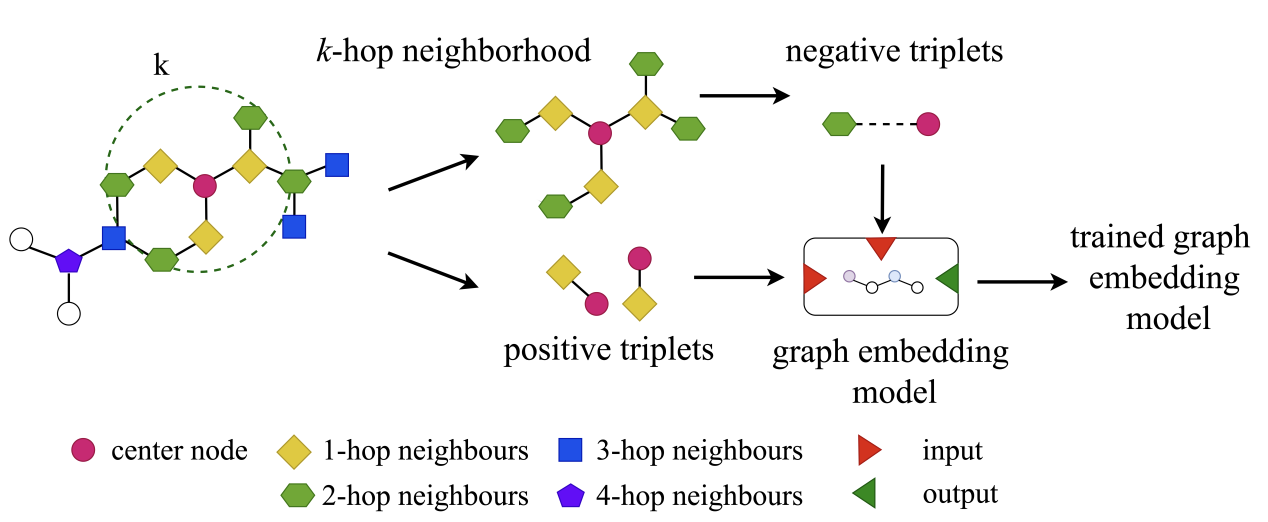}
  \vspace{-0.3cm}
  \caption{Integrating the structure of knowledge graph into negative sampling. Taken from ~\cite{ahrabian2020structure}.} 
  \label{fig: ns_in_kg}
  \vspace{-0.4cm}
\end{figure}

Negative sampling is a fundamental technique in knowledge graph embedding, which is applied to sample entities from the knowledge graph to replace the head entity or tail entity for forming a negative triple. A common negative sampling strategy in KGE is randomly sampling entities from a uniform distribution. However, such a strategy owns the obvious limitation that the sampled entities usually do not match the relations with the remaining entities, which does not provide meaningful information for gradient and prevents the model from learning better embeddings. To constrain the correlation of negative triples, ~\cite{krompass2015type} drew negative triples within the range constraints of entity types.
PNS~\cite{kanojia2017enhancing} proposed a probabilistic negative sampling to address the skewness issue in the dataset, which leveraged a tuning parameter to sample negatives from a pre-designed list that contained semantic possible negative instances.  
~\cite{kotnis2017analysis} developed two embedding-based negative sampling strategies: nearest neighbor sampling and near miss sampling, which aimed to search for negative triples that are close to positive triples in embedding space. KBGAN~\cite{cai2017kbgan} developed an adversarial framework for knowledge graph embeddings, which utilized existing embedding models as the generator to generate high-quality negative triples. Later, several works~\cite{wang2018incorporating, sun2019rotate,cai2017kbgan} continued to integrate GAN into knowledge graph embedding for sampling negative triples. For example, IGAN~\cite{wang2018incorporating} leveraged two-layer fully-connected neural networks as the generator to search informative negative triples. These methods sample high-quality negative triples with large gradients, which avoid the vanishing gradient issue and achieve performance improvements. However, GAN-based negative sampling methods need an extra generator and a complex gradient update approach. 
To efficiently sample negative triples, RotatE~\cite{sun2019rotate} proposed a self-adversarial negative sampling strategy that selected negatives based on the current model. Similar to DNS, Shan et al.~\cite{shan2018confidence} proposed a confidence-aware negative sampling method to enhance the confidence-aware knowledge representation learning (CKRL), which sampled negatives according to the softmax function that calculated by the current model on the candidate negatives set. SNS~\cite{islam2021simple} utilized a distance-based score to search for high-quality negatives from a small randomly sampled candidate negative set in embedding space. NSCaching~\cite{zhang2019nscaching} used extra caches to store large-gradient negative heads and tails for each positive triplet respectively, and directly sampled negatives from the caches. To sample valid negative triples for assist model training,  ANS~\cite{qin2019knowledge} leveraged the K-Means clustering algorithm to measure the similarity in embedding space and uniformly sampled negatives from the same cluster for a particular positive triplet. SANS~\cite{ahrabian2020structure} utilized knowledge graph structure to sample negative triples where 1-hop neighbors are regarded as positives and the $k$-hop neighbors $(k>1)$ serve as negatives (See Figure~\ref{fig: ns_in_kg}). Instead of selecting an existing entity for negative sampling, MixKG~\cite{che2022mixkg} leveraged a mixing operation to synthesize hard negative samples. 
Different from the abovementioned methods of replacing head or tail entities, TransG~\cite{xiao2015transg} constructed negative triples by substituting the relation of triples. To sum up, negative sampling in a knowledge graph should focus on the validity of negatives satisfied the semantic relation, which can provide more information for the model to distinguish between positives and negatives.

In summary, negative sampling in knowledge graph embedding relies heavily on the triple loss function. Thus, how to design a powerful loss function that incorporates the characteristics of a knowledge graph? How to develop a hierarchical negative sampling method by leveraging the hierarchical structure of the knowledge graph?

\subsection{Negative Sampling in NLP}
Negative sampling plays an important role in the field of natural language processing (NLP), which is widely used across a spectrum of applications including word embedding, sentence embedding, dialogue systems, dense retrieval (DR), and named entity recognition (NER). Word2vec sampled negative words according to the distribution of word frequency.
Goldberg et al.\cite{goldberg2014word2vec} attempted to explain negative sampling in word2vec. To alleviate the gradient vanishing issue in the skip-gram model with word-frequency negative sampling, Chen et al.~\cite{chen2018improving} dynamically selected informative negative samples based on self-embedded features. Rao et al.~\cite{rao2016noise} proposed three negative sampling strategies for negative answer selection, comprising random sampling, max sampling, and mix sampling. These sampling methods also be applied to open-domain dialogue systems for negative response selection~\cite{li2019sampling}. Besides, Li et al.~\cite{li2019sampling} also proposed a semi-hard sampling method where negatives satisfied a margin constraint. Negative sampling is an effective technique for named entity recognition (NER) models, which suffer greatly from unlabeled entity problems. Li et al.~\cite{li2020empirical} leveraged negative sampling to randomly sample a small subset of unlabeled instances rather than the whole set. After that, Li et al.~\cite{li2022rethinking} designed a weighted sampling distribution to replace random sampling for boosting performance.

In sentence embedding learning~\cite{guo2018effective, gao2021simcse, wu2020clear}, negative sampling plays a vital role in enhancing model performance by facilitating the distinction between semantically similar and dissimilar sentences. This technique has evolved from basic random sampling to more sophisticated strategies. Guo et al.~\cite{guo2018effective} proposed a hard negative mining method for bilingual sentence embedding learning, in which selected examples are close to the positive translation in semantic embedding space. Recently, contrastive learning for sentence embedding learning has achieved tremendous progress. CLEAR~\cite{wu2020clear} proposed sentence-level data augmentation for contrastive sentence representation learning, including word deletion, reordering, and substitution. SimCSE~\cite{gao2021simcse} proposed a model-level data augmentation that passed the same sentence twice with different dropout probabilities. The abovementioned methods focus on the sampling method for positive pairs where negatives come from the current mini-batch. CLINE~\cite{wang2021cline} selected semantic negative examples in embedding space. MoCoSE~\cite{cao2022exploring} leveraged negative sample queue to obtain better performance. VaSCL~\cite{zhang2021virtual} proposed neighborhood constrained contrastive learning, which utilized KNN to obtain top-K similar negatives from the current batch as hard negatives. SNCSE~\cite{wang2022sncse} utilized soft negative samples to enhance unsupervised sentence embedding learning where soft negatives were defined as the negation of original sentences with similar textual but different semantics. MixCSE~\cite{zhang2022unsupervised} adopted a mixing operation to synthesize hard negatives for unsupervised sentence representation learning.

In recent years, dense retrieval (DR) models have become a dominant technique to solve the semantic match problem~\cite{gillick2019learning, qu2020rocketqa, karpukhin2020dense}. Negative sampling is an indispensable component for training DR models. EBR~\cite{huang2020embedding} adopted a random sampling strategy to select negatives for embedding-based retrieval models, which leveraged the triplet loss for the recall optimization task. To improve the efficiency with the demand of a large number of negatives, several works~\cite{ gillick2019learning, karpukhin2020dense, qu2020rocketqa} employed In-batch NS where other examples in the same mini-batch are treated as negatives. In fact, In-batch NS is approximately equivalent to random negative sampling. Moreover, the experimental results in RocketQA~\cite{qu2020rocketqa} demonstrated that it is beneficial to increase the number of negatives by introducing cross-batch negatives. Another popular direction is to apply hard negatives to train DR models. Gillick et al.~\cite{gillick2019learning} selected the most similar 10 entities based on the current model as hard negatives. Karpukhin et al.~\cite{karpukhin2020dense} utilized top passages generated by BM25 as hard negatives. Xiong et al.~\cite{xiong2020approximate} proposed Approximate nearest neighbor Negative Contrastive Learning (ANCE) to select hard negatives by an ANN index. Different from static hard negative sampling methods, Zhan et al.~\cite{zhan2021optimizing} designed a dynamic hard negative sampling method, which utilized a trainable query encoder to retrieve the most challenging documents as hard negatives.

In summary, negative sampling in natural language processing still leaves a lot of room for exploration. For example, how to develop a better In-batch NS to mine hard negatives and simultaneously mitigate the false negative issue?

\subsection{Negative Sampling in CV} 
After several decades of sustained effort, a large number of supervised methods have achieved significant improvements in computer vision (CV), which leveraged large datasets with labeled examples to learn visual representations. However, large-scale datasets with labels require human annotation, which is very expensive and hurts applications on the Internet scale. Contrastive learning for unsupervised visual representation learning becomes a natural way to address this issue, which aims to learn visual embeddings on unlabeled data. 
Recent developments~\cite{wu2018unsupervised, oord2018representation, tian2020contrastive, he2020momentum, chen2020simple} in unsupervised visual representation learning present a promising potential by using contrastive loss, which aims to contrastive positive pairs and negative pairs. Besides, metric learning is one of the basic learning ways for computer vision.

Generally, negative examples can be obtained either within a mini-batch or from a memory bank. In-batch negative sampling~\cite{chen2020simple, oord2018representation} samples negatives from the current mini-batch. Memory-based negative sampling~\cite{wu2018unsupervised, misra2020self, tian2020contrastive} samples negatives from a memory bank that stores mini-batch samples from previous batches. MoCo~\cite{he2020momentum} argued that larger batch sizes play a significant role in model learning and maintained a queue to accumulate a large number of features that serve as negative samples for model training. SimCLR~\cite{chen2020simple} proposed a simple framework for contrastive learning, which used all other images in the current batch as negative samples. Without the assistance of labels, the false negative issue is an inevitable problem in contrastive learning, which sampled true label examples from the data distribution. To address this, Chuang et al.~\cite{chuang2020debiased} proposed a debiased contrastive loss (DCL) that corrected the weights of negatives in the objective. After that, Huynh et al.~\cite{huynh2022boosting} proposed false negative cancellation strategies consisting of elimination and attraction to improve contrastive learning. Besides, Cai et al.~\cite{cai2020all} conducted an empirical study to analyze the importance of negative samples and concluded that only 5\% hardest negatives are necessary for high-accuracy contrastive learning. Wang et al.~\cite{wang2021understanding} investigated the behavior of contrastive loss and concluded that loss optimization automatically focuses on hard negative samples which can be controlled by the temperature. To improve the performance and efficiency of contrastive learning, most strategies focus on hard negative sampling. MoCHi~\cite{kalantidis2020hard} proposed hard negative mixing strategies for contrastive learning to boost model learning, which synthesized hard negatives by mixing the hardest negatives and a positive query in the embedding space. Wu et al.~\cite{wu2020conditional} proposed conditional negative sampling to sample negatives within a ring where negatives in the ring are close but not too close to the positive example. Such a method is similar to the semi-hard negative sampling method. Similarly, Xie et al.~\cite{xie2020delving} proposed four negative sampling strategies based on the cosine distance criterion between the anchor and negative candidates, comprised of hard, semi-hard, random, and semi-easy. HCL~\cite{robinson2020contrastive} proposed an efficient hard negative sampling method that rewrote the important weights of each negative example, which assigned higher weights to negatives that are close to the anchor. Different from hard negative sampling that only selects hard negatives, HCL adopted DCL to alleviate the false negative issue. Ge et al.~\cite{ge2021robust} designed texture-based and patch-based negative sampling strategies to generate hard negative from the input images, which motivated the model to learn more semantics rather than superficial features. In addition, several works~\cite{ho2020contrastive, hu2021adco} leveraged adversarial learning to improve contrastive learning. For example, CLAE~\cite{ho2020contrastive} utilized adversarial examples and adversarial training for contrastive learning to generate hard negative examples. AdCo~\cite{hu2021adco} adopted GAN for contrastive learning for sampling more informative negative samples. In addition, time-contrastive learning~\cite{sermanet2016unsupervised,sermanet2018time,nair2022r3m} is particularly applied to video data where positive and negative examples for learning are drawn from different timestamps within video sequences. In this framework, the same time step across different camera views is similar (positive samples), while frames from different time steps are dissimilar (negative samples).

\begin{figure}[t]
  \centering
  \includegraphics[width=0.4\textwidth]{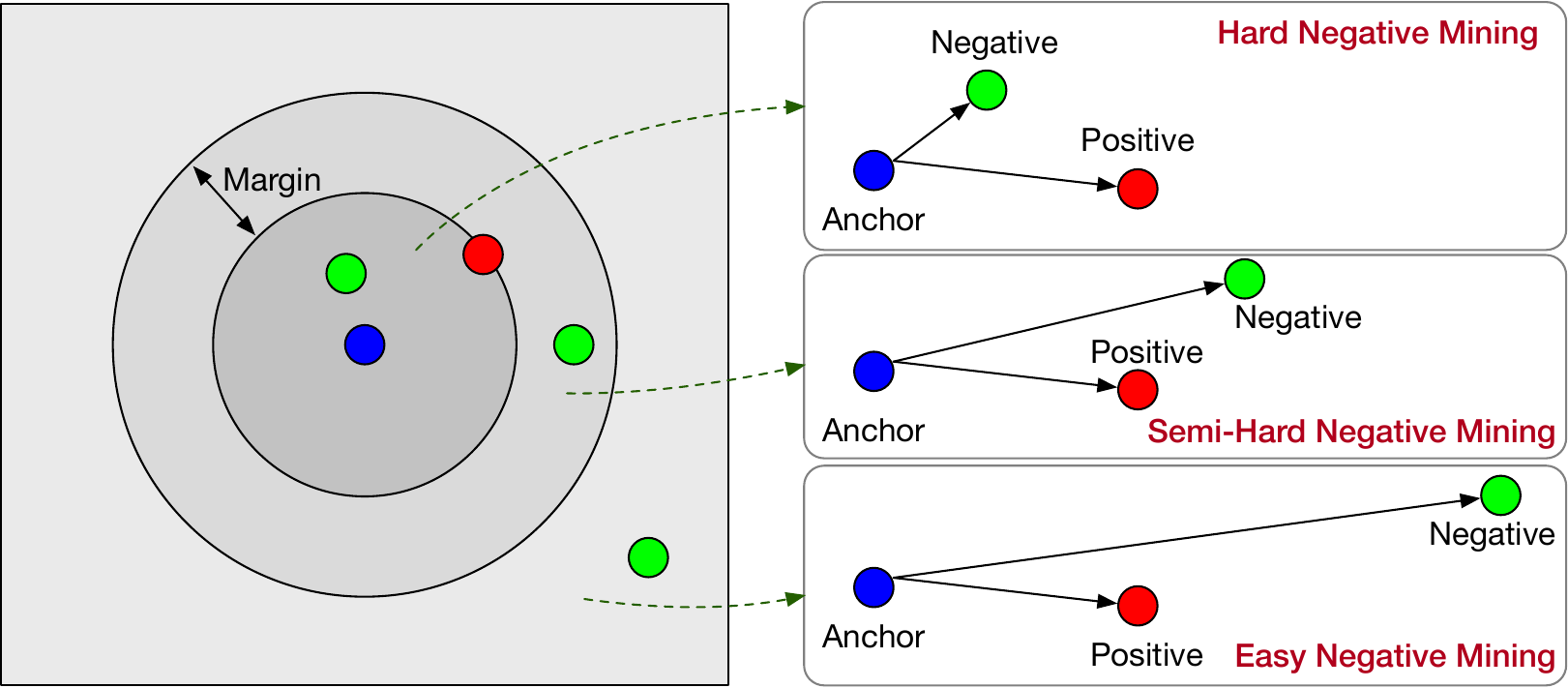}
  \vspace{-0.3cm}
  \caption{The selection of negative examples in computer vision based on metric learning. }  
  \label{fig: ns_in_metric}
  \vspace{-0.4cm}
\end{figure}

Furthermore, numerous methods~\cite{schroff2015facenet, wang2015unsupervised, shrivastava2016training} in computer vision are designed on the triple loss based on metric learning by leveraging a max-margin approach to distinguish positive pairs from negative pairs, which are widely used in object detection, image classification, and face recognition. The easiest negative sampling method is random sampling. However, most randomly sampled negative examples are easy examples that easily satisfy the margin constraint, which contributes less to the gradients. To address this problem, \cite{simo2015discriminative} focused on mining hard negative examples that are close in embedding space to enhance the learning process. Wang et al.~\cite{wang2015unsupervised} mined hard negative triples after 10 epochs of training with randomly sampled negatives. However, mining hard negatives requires searching the whole training set, which is computationally expensive. Besides, the results in FaceNet~\cite{schroff2015facenet} showed that the hardest negative examples significantly decrease the convergence speed and proposed a semi-hard negative mining strategy. To reduce the computational complexity, Mao et al.~\cite{mao2019metric} sampled semi-hard negative examples from a mini-batch rather than the entire training set. Wu et al.~\cite{wu2017sampling} proposed a distance-weighted negative sampling strategy to optimize the triple loss. Iscen et al.~\cite{iscen2018mining} mined hard negatives for an anchor from its nearest Euclidean neighbors rather than manifold neighbors defined over the Euclidean nearest neighbor graph.  
HDC~\cite{yuan2017hard} proposed a hard-aware deeply cascaded embedding to sample negatives from different hard levels of samples. Harwood et al.~\cite{harwood2017smart} proposed a smart negative sampling strategy for deep metric learning to sample a smart negative with a tuning variable and approximate positive examples. Although hard negatives provide large gradients for model optimization, are easy negatives really useless for metric learning? DAML~\cite{duan2018deep} argued that easy negatives should not be neglected since these play a supplementary role in hard negatives, which utilized a hard negative generator to synthesize hard negatives from easy ones. 
Here, we plot a figure to intuitively illustrate the selection of negative examples in metric learning (See Figure~\ref{fig: ns_in_metric}). Hard negative mining samples negatives that are too close to the anchor, which leads to a high variance on the gradient and hinders the model from learning better representations. Semi-hard negative mining selects negatives that are challenging yet not overwhelmingly so, promising the similarity of the positive pair is higher than the negatives.

In summary, negative sampling in computer vision has left us a lot of room to explore. For example, how many negative samples are the best choices? Due to the emergence of negative-free methods, are negative samples really needed?

\label{conclusion}
\section{Discussion and Future Directions}\label{sec:discussion}
In this section, we discuss several open problems in negative sampling and provide future directions for negative sampling to facilitate the development of this field. Is negative sampling necessary? If not, what kind of training paradigm is needed, and if so, how many negative samples are needed, and what quality of negative samples are needed?

\noindent \textbf{Non-Sampling.} Despite a myriad of negative sampling methods that have emerged recently, another hot direction is a non-sampling strategy that takes all negative samples into consideration in model optimization. Non-sampling strategy generally assigns lower weights for negative samples compared to positive ones. Such a setting is consistent with our intuition that positive samples should be evaluated with higher weights than negative ones. Due to the efficiency of non-sampling strategy, several efforts~\cite{xin2018batch, chen2019efficient, chen2020efficient, chen2020jointly} focus on promoting the efficiency of mini-batch Stochastic Gradient Descent (SGD) based methods. For example, Chen et al.~\cite{chen2019efficient} adopted non-sampling strategies for recommendation.
Such a non-sampling strategy is also applicable for knowledge graph embedding learning~\cite{chen2020jointly} and word embedding~\cite{xin2018batch}. Although the non-sampling strategy can provide a more stable way for model optimization, its efficiency is a non-negligible issue. How to design a more efficient method for model optimization incorporated with a non-sampling strategy?  Furthermore, a non-sampling strategy for other domains is not fully explored.

\noindent \textbf{Getting Rid of Negative Sampling.} Due to the quality and quantity of negative samples playing a significant impact on downstream performance, several works~\cite{caron2020unsupervised,grill2020bootstrap,chen2021exploring,thakoor2021large} attempt to get rid of negative sampling and adopt other paradigms for model learning. SwAV~\cite{caron2020unsupervised} adopted online clustering to compare the consistency between cluster assignments rather than image features, which did not require explicit negative instances for unsupervised visual learning. BYOL~\cite{grill2020bootstrap} only utilized positive pairs without negative pairs for self-supervised learning, which adopted two neural networks to directly achieve prediction from one view to another view for the same image. SimSiam~\cite{chen2021exploring} presented a simple siamese network for representation learning, which also discards negative sample pairs.  BGRL~\cite{thakoor2021large} also alleviates negative sampling for large-scale graph representation learning. Furthermore, generative self-supervised learning methods without negative sampling have been successfully applied in NLP~\cite{devlin2018bert}, CV~\cite{he2022masked} and graph~\cite{hou2022graphmae}. Thus, a promising research direction is to explore new learning methods or other alternatives to negative sampling.

\noindent \textbf{The Quantity of Negative Samples.} How many negative samples are needed? 
Arora et al.~\cite{arora2019theoretical} proposed a theoretical analysis for contrastive learning to demonstrate a performance degradation by using larger negative samples. Wu et al.~\cite{wu2021rethinking} 
proposed an adaptive negative sampling (ANS) method to dynamically adjust the ratio during the training process. Ash et al.~\cite{ash2021investigating} continued to investigate the number of negative examples in contrastive learning and revealed that the selection of optimal negative example size relies on the underlying concepts in the data. 
To address this gap between the theoretical analysis and empirical results, Nozawa et al.~\cite{nozawa2021understanding} proposed a lower bound for self-supervised learning, which aimed to adjust collision probability according to the number of negative samples. Recently, Awasthi et al.~\cite{awasthi2022more} argued that a collision-coverage trade-off is not an inherent property in contrastive learning, and claimed that the downstream performance does not degrade with the increasing of negative examples in a simple theoretical setting. Furthermore, Sohn~\cite{sohn2016improved} proposed an N-pair loss by adopting multiple negative samples for boosting deep metric learning. Although some works attempt to explain or explore the impact of negative sample size, there is still no one answer to what is the standard of negative sample size for a specific domain or even task. In practical applications, the quantity of negative samples is determined by a large number of trials. In the future, we are eager to obtain a criterion for a negative sample size.

\noindent \textbf{The Quality of Negative Samples.} Are the hardest negative samples the best ones? Does model optimization really not require easy negative samples? Results in several efforts demonstrated that easy negative samples play a crucial role in the early training stage. For example, Wang et al.~\cite{wang2015unsupervised} conducted random sampling to mine easy negatives in the first 10 epochs before mining hard negatives. CuCo~\cite{chu2021cuco} utilized curriculum learning to mine negative samples from easy to hard. DAML~\cite{duan2018deep} highlighted the importance of easy negative samples for early stable training. Furthermore, many works have demonstrated that the hardest negative samples are not beneficial for model stability and robustness. For example, FaceNet~\cite{schroff2015facenet} proposed semi-hard negative samples. Wu et al.~\cite{wu2020conditional} proposed to sample negatives with a ring constraint. Therefore, hard negative sampling should fully consider easy negatives and control the hardness of hard negatives. How to naturally and dynamically add easy negatives into hard negative sampling is a vital and worth exploring problem. Hardness of negative samples for model optimization is also a valuable research direction.

\noindent \textbf{False Negative Issue.} As a common and inevitable challenge for negative sampling, the false negative issue mainly comes from two aspects. The first is derived from unlabeled data in contrastive learning. 
Negative sampling in contrastive learning simultaneously focuses on mining hard negatives and mitigating the false negative issue. The second is that false negative instances naturally exist in implicit feedback in the recommendation domain. 
Although some strategies are employed to alleviate the false negative issue, it is impossible to eliminate this issue fundamentally under the paradigm of contrastive learning. Generative self-supervised learning can completely avoid this issue, which gets rid of negative sampling and becomes a powerful alternative for contrastive learning in unsupervised representation learning. In summary, alleviating or even eliminating the false negative issue is a meaningful research direction.

\section{Conclusion}\label{sec:conclusion}
In this survey, we have conducted an extensive review of the landscape of negative sampling techniques across a multitude of domains. Negative sampling is a fundamental technique in machine learning, which can accelerate the training process and boost downstream performance. We summarize a general negative sampling framework and develop a tool that contains many negative sampling methods among various domains. The selection methods for negative candidates are summarized, including global, local, mini-batch, hop, and memory-based. Besides, we categorize all existing negative sampling methods into five groups (static, hard, GAN-based, Auxiliary-based, and In-batch) and demonstrate their pros and cons. Furthermore, we illustrate negative sampling applications in various domains. Finally, open problems and future directions of negative sampling are presented.

  \section*{Acknowledgments}


This work is supported by the National Key Research and Development Program of China 2021YFF1201300, the NSFC for Distinguished Young Scholar 61825602 and the New Cornerstone Science Foundation through the XPLORER PRIZE.



\ifCLASSOPTIONcaptionsoff
  \newpage
\fi


\bibliographystyle{IEEEtran}
\bibliography{IEEEabrv,mybib}

\begin{thebibliography}{100}
\providecommand{\url}[1]{#1}
\csname url@samestyle\endcsname
\providecommand{\newblock}{\relax}
\providecommand{\bibinfo}[2]{#2}
\providecommand{\BIBentrySTDinterwordspacing}{\spaceskip=0pt\relax}
\providecommand{\BIBentryALTinterwordstretchfactor}{4}
\providecommand{\BIBentryALTinterwordspacing}{\spaceskip=\fontdimen2\font plus
\BIBentryALTinterwordstretchfactor\fontdimen3\font minus \fontdimen4\font\relax}
\providecommand{\BIBforeignlanguage}[2]{{%
\expandafter\ifx\csname l@#1\endcsname\relax
\typeout{** WARNING: IEEEtran.bst: No hyphenation pattern has been}%
\typeout{** loaded for the language `#1'. Using the pattern for}%
\typeout{** the default language instead.}%
\else
\language=\csname l@#1\endcsname
\fi
#2}}
\providecommand{\BIBdecl}{\relax}
\BIBdecl

\bibitem{mikolov2013distributed}
T.~Mikolov, I.~Sutskever, K.~Chen, G.~S. Corrado, and J.~Dean, ``Distributed representations of words and phrases and their compositionality,'' \emph{NIPS}, vol.~26, 2013.

\bibitem{rendle2012bpr}
S.~Rendle, C.~Freudenthaler, Z.~Gantner, and L.~Schmidt-Thieme, ``Bpr: Bayesian personalized ranking from implicit feedback,'' \emph{arXiv preprint arXiv:1205.2618}, 2012.

\bibitem{perozzi2014deepwalk}
B.~Perozzi, R.~Al-Rfou, and S.~Skiena, ``Deepwalk: Online learning of social representations,'' in \emph{KDD}, 2014, pp. 701--710.

\bibitem{tang2015line}
J.~Tang, M.~Qu, M.~Wang, M.~Zhang, J.~Yan, and Q.~Mei, ``Line: Large-scale information network embedding,'' in \emph{WWW}, 2015, pp. 1067--1077.

\bibitem{grover2016node2vec}
A.~Grover and J.~Leskovec, ``node2vec: Scalable feature learning for networks,'' in \emph{KDD}, 2016, pp. 855--864.

\bibitem{yang2020understanding}
Z.~Yang, M.~Ding, C.~Zhou, H.~Yang, J.~Zhou, and J.~Tang, ``Understanding negative sampling in graph representation learning,'' in \emph{KDD}, 2020, pp. 1666--1676.

\bibitem{huang2021mixgcf}
T.~Huang, Y.~Dong, M.~Ding, Z.~Yang, W.~Feng, X.~Wang, and J.~Tang, ``Mixgcf: An improved training method for graph neural network-based recommender systems,'' in \emph{KDD}, 2021, pp. 665--674.

\bibitem{krompass2015type}
D.~Krompa{\ss}, S.~Baier, and V.~Tresp, ``Type-constrained representation learning in knowledge graphs,'' in \emph{International semantic web conference}.\hskip 1em plus 0.5em minus 0.4em\relax Springer, 2015, pp. 640--655.

\bibitem{cai2017kbgan}
L.~Cai and W.~Y. Wang, ``Kbgan: Adversarial learning for knowledge graph embeddings,'' \emph{arXiv preprint arXiv:1711.04071}, 2017.

\bibitem{grbovic2015context}
M.~Grbovic, N.~Djuric, V.~Radosavljevic, F.~Silvestri, and N.~Bhamidipati, ``Context-and content-aware embeddings for query rewriting in sponsored search,'' in \emph{SIGIR}, 2015, pp. 383--392.

\bibitem{zhang2018gneg}
Z.~Zhang and P.~Zweigenbaum, ``Gneg: Graph-based negative sampling for word2vec,'' in \emph{ACL}, 2018, pp. 566--571.

\bibitem{wu2021esimcse}
X.~Wu, C.~Gao, L.~Zang, J.~Han, Z.~Wang, and S.~Hu, ``Esimcse: Enhanced sample building method for contrastive learning of unsupervised sentence embedding,'' \emph{arXiv preprint arXiv:2109.04380}, 2021.

\bibitem{schroff2015facenet}
F.~Schroff, D.~Kalenichenko, and J.~Philbin, ``Facenet: A unified embedding for face recognition and clustering,'' in \emph{CVPR}, 2015, pp. 815--823.

\bibitem{wu2017sampling}
C.-Y. Wu, R.~Manmatha, A.~J. Smola, and P.~Krahenbuhl, ``Sampling matters in deep embedding learning,'' in \emph{ICCV}, 2017, pp. 2840--2848.

\bibitem{chuang2020debiased}
C.-Y. Chuang, J.~Robinson, Y.-C. Lin, A.~Torralba, and S.~Jegelka, ``Debiased contrastive learning,'' \emph{NIPS}, vol.~33, pp. 8765--8775, 2020.

\bibitem{kalantidis2020hard}
Y.~Kalantidis, M.~B. Sariyildiz, N.~Pion, P.~Weinzaepfel, and D.~Larlus, ``Hard negative mixing for contrastive learning,'' \emph{NIPS}, vol.~33, pp. 21\,798--21\,809, 2020.

\bibitem{robinson2020contrastive}
J.~Robinson, C.-Y. Chuang, S.~Sra, and S.~Jegelka, ``Contrastive learning with hard negative samples,'' \emph{arXiv preprint arXiv:2010.04592}, 2020.

\bibitem{wu2020conditional}
M.~Wu, M.~Mosse, C.~Zhuang, D.~Yamins, and N.~Goodman, ``Conditional negative sampling for contrastive learning of visual representations,'' \emph{arXiv preprint arXiv:2010.02037}, 2020.

\bibitem{zhang2013optimizing}
W.~Zhang, T.~Chen, J.~Wang, and Y.~Yu, ``Optimizing top-n collaborative filtering via dynamic negative item sampling,'' in \emph{SIGIR}, 2013, pp. 785--788.

\bibitem{bordes2013translating}
A.~Bordes, N.~Usunier, A.~Garcia-Duran, J.~Weston, and O.~Yakhnenko, ``Translating embeddings for modeling multi-relational data,'' \emph{NIPS}, vol.~26, 2013.

\bibitem{kipf2019contrastive}
T.~Kipf, E.~Van~der Pol, and M.~Welling, ``Contrastive learning of structured world models,'' \emph{arXiv preprint arXiv:1911.12247}, 2019.

\bibitem{kipf2016semi}
T.~N. Kipf and M.~Welling, ``Semi-supervised classification with graph convolutional networks,'' \emph{arXiv preprint arXiv:1609.02907}, 2016.

\bibitem{velivckovic2017graph}
P.~Veli{\v{c}}kovi{\'c}, G.~Cucurull, A.~Casanova, A.~Romero, P.~Lio, and Y.~Bengio, ``Graph attention networks,'' \emph{arXiv preprint arXiv:1710.10903}, 2017.

\bibitem{hamilton2017inductive}
W.~Hamilton, Z.~Ying, and J.~Leskovec, ``Inductive representation learning on large graphs,'' \emph{NIPS}, vol.~30, 2017.

\bibitem{he2016deep}
K.~He, X.~Zhang, S.~Ren, and J.~Sun, ``Deep residual learning for image recognition,'' in \emph{CVPR}, 2016, pp. 770--778.

\bibitem{devlin2018bert}
J.~Devlin, M.-W. Chang, K.~Lee, and K.~Toutanova, ``Bert: Pre-training of deep bidirectional transformers for language understanding,'' \emph{arXiv preprint arXiv:1810.04805}, 2018.

\bibitem{liu2019roberta}
Y.~Liu, M.~Ott, N.~Goyal, J.~Du, M.~Joshi, D.~Chen, O.~Levy, M.~Lewis, L.~Zettlemoyer, and V.~Stoyanov, ``Roberta: A robustly optimized bert pretraining approach,'' \emph{arXiv preprint arXiv:1907.11692}, 2019.

\bibitem{wang2014knowledge}
Z.~Wang, J.~Zhang, J.~Feng, and Z.~Chen, ``Knowledge graph embedding by translating on hyperplanes,'' in \emph{AAAI}, vol.~28, no.~1, 2014.

\bibitem{pan2008one}
R.~Pan, Y.~Zhou, B.~Cao, N.~N. Liu, R.~Lukose, M.~Scholz, and Q.~Yang, ``One-class collaborative filtering,'' in \emph{ICDM}.\hskip 1em plus 0.5em minus 0.4em\relax IEEE, 2008, pp. 502--511.

\bibitem{xiao2015transg}
H.~Xiao, M.~Huang, Y.~Hao, and X.~Zhu, ``Transg: A generative mixture model for knowledge graph embedding,'' \emph{arXiv preprint arXiv:1509.05488}, 2015.

\bibitem{wang2019neural}
X.~Wang, X.~He, M.~Wang, F.~Feng, and T.-S. Chua, ``Neural graph collaborative filtering,'' in \emph{SIGIR}, 2019, pp. 165--174.

\bibitem{he2020lightgcn}
X.~He, K.~Deng, X.~Wang, Y.~Li, Y.~Zhang, and M.~Wang, ``Lightgcn: Simplifying and powering graph convolution network for recommendation,'' in \emph{SIGIR}, 2020, pp. 639--648.

\bibitem{dalal2005histograms}
N.~Dalal and B.~Triggs, ``Histograms of oriented gradients for human detection,'' in \emph{CVPR}, vol.~1.\hskip 1em plus 0.5em minus 0.4em\relax Ieee, 2005, pp. 886--893.

\bibitem{felzenszwalb2009object}
P.~F. Felzenszwalb, R.~B. Girshick, D.~McAllester, and D.~Ramanan, ``Object detection with discriminatively trained part-based models,'' \emph{IEEE transactions on pattern analysis and machine intelligence}, vol.~32, no.~9, pp. 1627--1645, 2009.

\bibitem{malisiewicz2011ensemble}
T.~Malisiewicz, A.~Gupta, and A.~A. Efros, ``Ensemble of exemplar-svms for object detection and beyond,'' in \emph{2011 International conference on computer vision}.\hskip 1em plus 0.5em minus 0.4em\relax IEEE, 2011, pp. 89--96.

\bibitem{chen2018improving}
L.~Chen, F.~Yuan, J.~M. Jose, and W.~Zhang, ``Improving negative sampling for word representation using self-embedded features,'' in \emph{WSDM}, 2018, pp. 99--107.

\bibitem{rao2016noise}
J.~Rao, H.~He, and J.~Lin, ``Noise-contrastive estimation for answer selection with deep neural networks,'' in \emph{CIKM}, 2016, pp. 1913--1916.

\bibitem{sun2018bootstrapping}
Z.~Sun, W.~Hu, Q.~Zhang, and Y.~Qu, ``Bootstrapping entity alignment with knowledge graph embedding.'' in \emph{IJCAI}, vol.~18, 2018, pp. 4396--4402.

\bibitem{ying2018graph}
R.~Ying, R.~He, K.~Chen, P.~Eksombatchai, W.~L. Hamilton, and J.~Leskovec, ``Graph convolutional neural networks for web-scale recommender systems,'' in \emph{KDD}, 2018, pp. 974--983.

\bibitem{wang2020reinforced}
X.~Wang, Y.~Xu, X.~He, Y.~Cao, M.~Wang, and T.-S. Chua, ``Reinforced negative sampling over knowledge graph for recommendation,'' in \emph{WWW}, 2020, pp. 99--109.

\bibitem{shrivastava2016training}
A.~Shrivastava, A.~Gupta, and R.~Girshick, ``Training region-based object detectors with online hard example mining,'' in \emph{CVPR}, 2016, pp. 761--769.

\bibitem{goodfellow2014generative}
I.~Goodfellow, J.~Pouget-Abadie, M.~Mirza, B.~Xu, D.~Warde-Farley, S.~Ozair, A.~Courville, and Y.~Bengio, ``Generative adversarial nets,'' \emph{NIPS}, vol.~27, 2014.

\bibitem{wang2017irgan}
J.~Wang, L.~Yu, W.~Zhang, Y.~Gong, Y.~Xu, B.~Wang, P.~Zhang, and D.~Zhang, ``Irgan: A minimax game for unifying generative and discriminative information retrieval models,'' in \emph{SIGIR}, 2017, pp. 515--524.

\bibitem{wang2019learning}
H.~Wang, J.~Wang, J.~Wang, M.~Zhao, W.~Zhang, F.~Zhang, W.~Li, X.~Xie, and M.~Guo, ``Learning graph representation with generative adversarial nets,'' \emph{TKDE}, vol.~33, no.~8, pp. 3090--3103, 2019.

\bibitem{wang2018neural}
Q.~Wang, H.~Yin, Z.~Hu, D.~Lian, H.~Wang, and Z.~Huang, ``Neural memory streaming recommender networks with adversarial training,'' in \emph{KDD}, 2018, pp. 2467--2475.

\bibitem{chae2018cfgan}
D.-K. Chae, J.-S. Kang, S.-W. Kim, and J.-T. Lee, ``Cfgan: A generic collaborative filtering framework based on generative adversarial networks,'' in \emph{CIKM}, 2018, pp. 137--146.

\bibitem{park2019adversarial}
D.~H. Park and Y.~Chang, ``Adversarial sampling and training for semi-supervised information retrieval,'' in \emph{WWW}, 2019, pp. 1443--1453.

\bibitem{chen2020simple}
T.~Chen, S.~Kornblith, M.~Norouzi, and G.~Hinton, ``A simple framework for contrastive learning of visual representations,'' in \emph{ICML}.\hskip 1em plus 0.5em minus 0.4em\relax PMLR, 2020, pp. 1597--1607.

\bibitem{he2020momentum}
K.~He, H.~Fan, Y.~Wu, S.~Xie, and R.~Girshick, ``Momentum contrast for unsupervised visual representation learning,'' in \emph{CVPR}, 2020, pp. 9729--9738.

\bibitem{hu2021adco}
Q.~Hu, X.~Wang, W.~Hu, and G.-J. Qi, ``Adco: Adversarial contrast for efficient learning of unsupervised representations from self-trained negative adversaries,'' in \emph{CVPR}, 2021, pp. 1074--1083.

\bibitem{xie2020delving}
J.~Xie, X.~Zhan, Z.~Liu, Y.~S. Ong, and C.~C. Loy, ``Delving into inter-image invariance for unsupervised visual representations,'' \emph{arXiv preprint arXiv:2008.11702}, 2020.

\bibitem{yang2023batchsampler}
Z.~Yang, T.~Huang, M.~Ding, Y.~Dong, R.~Ying, Y.~Cen, Y.~Geng, and J.~Tang, ``Batchsampler: Sampling mini-batches for contrastive learning in vision, language, and graphs,'' \emph{arXiv preprint arXiv:2306.03355}, 2023.

\bibitem{cai2020all}
T.~T. Cai, J.~Frankle, D.~J. Schwab, and A.~S. Morcos, ``Are all negatives created equal in contrastive instance discrimination?'' \emph{arXiv preprint arXiv:2010.06682}, 2020.

\bibitem{caselles2018word2vec}
H.~Caselles-Dupr{\'e}, F.~Lesaint, and J.~Royo-Letelier, ``Word2vec applied to recommendation: Hyperparameters matter,'' in \emph{RecSys}, 2018, pp. 352--356.

\bibitem{yang2014embedding}
B.~Yang, W.-t. Yih, X.~He, J.~Gao, and L.~Deng, ``Embedding entities and relations for learning and inference in knowledge bases,'' \emph{arXiv preprint arXiv:1412.6575}, 2014.

\bibitem{tao2018ruber}
C.~Tao, L.~Mou, D.~Zhao, and R.~Yan, ``Ruber: An unsupervised method for automatic evaluation of open-domain dialog systems,'' in \emph{AAAI}, 2018.

\bibitem{ghazarian2019better}
S.~Ghazarian, J.~T.-Z. Wei, A.~Galstyan, and N.~Peng, ``Better automatic evaluation of open-domain dialogue systems with contextualized embeddings,'' \emph{arXiv preprint arXiv:1904.10635}, 2019.

\bibitem{bucher2016hard}
M.~Bucher, S.~Herbin, and F.~Jurie, ``Hard negative mining for metric learning based zero-shot classification,'' in \emph{ECCV}.\hskip 1em plus 0.5em minus 0.4em\relax Springer, 2016, pp. 524--531.

\bibitem{harwood2017smart}
B.~Harwood, V.~Kumar~BG, G.~Carneiro, I.~Reid, and T.~Drummond, ``Smart mining for deep metric learning,'' in \emph{ICCV}, 2017, pp. 2821--2829.

\bibitem{galanopoulos2021hard}
D.~Galanopoulos and V.~Mezaris, ``Hard-negatives or non-negatives? a hard-negative selection strategy for cross-modal retrieval using the improved marginal ranking loss,'' in \emph{ICCV}, 2021, pp. 2312--2316.

\bibitem{rendle2014improving}
S.~Rendle and C.~Freudenthaler, ``Improving pairwise learning for item recommendation from implicit feedback,'' in \emph{WSDM}, 2014, pp. 273--282.

\bibitem{mao2021boosting}
X.~Mao, W.~Wang, Y.~Wu, and M.~Lan, ``Boosting the speed of entity alignment 10$\times$: Dual attention matching network with normalized hard sample mining,'' in \emph{WWW}, 2021, pp. 821--832.

\bibitem{zhang2019nscaching}
Y.~Zhang, Q.~Yao, Y.~Shao, and L.~Chen, ``Nscaching: simple and efficient negative sampling for knowledge graph embedding,'' in \emph{ICDE}.\hskip 1em plus 0.5em minus 0.4em\relax IEEE, 2019, pp. 614--625.

\bibitem{cao2022exploring}
R.~Cao, Y.~Wang, Y.~Liang, L.~Gao, J.~Zheng, J.~Ren, and Z.~Wang, ``Exploring the impact of negative samples of contrastive learning: A case study of sentence embedding,'' \emph{arXiv preprint arXiv:2202.13093}, 2022.

\bibitem{che2022mixkg}
F.~Che, G.~Yang, P.~Shao, D.~Zhang, and J.~Tao, ``Mixkg: Mixing for harder negative samples in knowledge graph,'' \emph{arXiv preprint arXiv:2202.09606}, 2022.

\bibitem{yu2017seqgan}
L.~Yu, W.~Zhang, J.~Wang, and Y.~Yu, ``Seqgan: Sequence generative adversarial nets with policy gradient,'' in \emph{AAAI}, vol.~31, no.~1, 2017.

\bibitem{bose2018adversarial}
A.~J. Bose, H.~Ling, and Y.~Cao, ``Adversarial contrastive estimation,'' \emph{arXiv preprint arXiv:1805.03642}, 2018.

\bibitem{wang2018incorporating}
P.~Wang, S.~Li, and R.~Pan, ``Incorporating gan for negative sampling in knowledge representation learning,'' in \emph{AAAI}, vol.~32, no.~1, 2018.

\bibitem{gao2019progan}
H.~Gao, J.~Pei, and H.~Huang, ``Progan: Network embedding via proximity generative adversarial network,'' in \emph{KDD}, 2019, pp. 1308--1316.

\bibitem{hu2019adversarial}
B.~Hu, Y.~Fang, and C.~Shi, ``Adversarial learning on heterogeneous information networks,'' in \emph{KDD}, 2019, pp. 120--129.

\bibitem{gupta2021synthesizing}
P.~Gupta, Y.~Tsvetkov, and J.~P. Bigham, ``Synthesizing adversarial negative responses for robust response ranking and evaluation,'' \emph{arXiv preprint arXiv:2106.05894}, 2021.

\bibitem{sinha2021negative}
A.~Sinha, K.~Ayush, J.~Song, B.~Uzkent, H.~Jin, and S.~Ermon, ``Negative data augmentation,'' \emph{arXiv preprint arXiv:2102.05113}, 2021.

\bibitem{ho2020contrastive}
C.-H. Ho and N.~Nvasconcelos, ``Contrastive learning with adversarial examples,'' \emph{NIPS}, vol.~33, pp. 17\,081--17\,093, 2020.

\bibitem{wang2021instance}
W.~Wang, W.~Zhou, J.~Bao, D.~Chen, and H.~Li, ``Instance-wise hard negative example generation for contrastive learning in unpaired image-to-image translation,'' in \emph{ICCV}, 2021, pp. 14\,020--14\,029.

\bibitem{duan2018deep}
Y.~Duan, W.~Zheng, X.~Lin, J.~Lu, and J.~Zhou, ``Deep adversarial metric learning,'' in \emph{CVPR}, 2018, pp. 2780--2789.

\bibitem{ahrabian2020structure}
K.~Ahrabian, A.~Feizi, Y.~Salehi, W.~L. Hamilton, and A.~J. Bose, ``Structure aware negative sampling in knowledge graphs,'' \emph{arXiv preprint arXiv:2009.11355}, 2020.

\bibitem{chen2019samwalker}
J.~Chen, C.~Wang, S.~Zhou, Q.~Shi, Y.~Feng, and C.~Chen, ``Samwalker: Social recommendation with informative sampling strategy,'' in \emph{WWW}, 2019, pp. 228--239.

\bibitem{wang2021dskreg}
Y.~Wang, Z.~Liu, Z.~Fan, L.~Sun, and P.~S. Yu, ``Dskreg: Differentiable sampling on knowledge graph for recommendation with relational gnn,'' in \emph{CIKM}, 2021, pp. 3513--3517.

\bibitem{zhao2014leveraging}
T.~Zhao, J.~McAuley, and I.~King, ``Leveraging social connections to improve personalized ranking for collaborative filtering,'' in \emph{CIKM}, 2014, pp. 261--270.

\bibitem{manotumruksa2017personalised}
J.~Manotumruksa, C.~Macdonald, and I.~Ounis, ``A personalised ranking framework with multiple sampling criteria for venue recommendation,'' in \emph{CIKM}, 2017, pp. 1469--1478.

\bibitem{loni2016bayesian}
B.~Loni, R.~Pagano, M.~Larson, and A.~Hanjalic, ``Bayesian personalized ranking with multi-channel user feedback,'' in \emph{RecSys}, 2016, pp. 361--364.

\bibitem{ding2018improved}
J.~Ding, F.~Feng, X.~He, G.~Yu, Y.~Li, and D.~Jin, ``An improved sampler for bayesian personalized ranking by leveraging view data,'' in \emph{WWW}, 2018, pp. 13--14.

\bibitem{ding2019reinforced}
J.~Ding, Y.~Quan, X.~He, Y.~Li, and D.~Jin, ``Reinforced negative sampling for recommendation with exposure data.'' in \emph{IJCAI}, 2019, pp. 2230--2236.

\bibitem{yang2022region}
Z.~Yang, M.~Ding, X.~Zou, J.~Tang, B.~Xu, C.~Zhou, and H.~Yang, ``Region or global a principle for negative sampling in graph-based recommendation,'' \emph{TKDE}, 2022.

\bibitem{wu2018unsupervised}
Z.~Wu, Y.~Xiong, S.~X. Yu, and D.~Lin, ``Unsupervised feature learning via non-parametric instance discrimination,'' in \emph{CVPR}, 2018, pp. 3733--3742.

\bibitem{ding2020simplify}
J.~Ding, Y.~Quan, Q.~Yao, Y.~Li, and D.~Jin, ``Simplify and robustify negative sampling for implicit collaborative filtering,'' \emph{NIPS}, vol.~33, pp. 1094--1105, 2020.

\bibitem{chen2020improved}
X.~Chen, H.~Fan, R.~Girshick, and K.~He, ``Improved baselines with momentum contrastive learning,'' \emph{arXiv preprint arXiv:2003.04297}, 2020.

\bibitem{qiu2020gcc}
J.~Qiu, Q.~Chen, Y.~Dong, J.~Zhang, H.~Yang, M.~Ding, K.~Wang, and J.~Tang, ``Gcc: Graph contrastive coding for graph neural network pre-training,'' in \emph{KDD}, 2020, pp. 1150--1160.

\bibitem{chen2017sampling}
T.~Chen, Y.~Sun, Y.~Shi, and L.~Hong, ``On sampling strategies for neural network-based collaborative filtering,'' in \emph{KDD}, 2017, pp. 767--776.

\bibitem{zhou2020s3}
K.~Zhou, H.~Wang, W.~X. Zhao, Y.~Zhu, S.~Wang, F.~Zhang, Z.~Wang, and J.-R. Wen, ``S3-rec: Self-supervised learning for sequential recommendation with mutual information maximization,'' in \emph{CIKM}, 2020, pp. 1893--1902.

\bibitem{wu2021self}
J.~Wu, X.~Wang, F.~Feng, X.~He, L.~Chen, J.~Lian, and X.~Xie, ``Self-supervised graph learning for recommendation,'' in \emph{SIGIR}, 2021, pp. 726--735.

\bibitem{yu2021self}
J.~Yu, H.~Yin, J.~Li, Q.~Wang, N.~Q.~V. Hung, and X.~Zhang, ``Self-supervised multi-channel hypergraph convolutional network for social recommendation,'' in \emph{WWW}, 2021, pp. 413--424.

\bibitem{xia2021self}
X.~Xia, H.~Yin, J.~Yu, Q.~Wang, L.~Cui, and X.~Zhang, ``Self-supervised hypergraph convolutional networks for session-based recommendation,'' in \emph{AAAI}, vol.~35, no.~5, 2021, pp. 4503--4511.

\bibitem{hassani2020contrastive}
K.~Hassani and A.~H. Khasahmadi, ``Contrastive multi-view representation learning on graphs,'' in \emph{ICML}.\hskip 1em plus 0.5em minus 0.4em\relax PMLR, 2020, pp. 4116--4126.

\bibitem{zhu2020deep}
Y.~Zhu, Y.~Xu, F.~Yu, Q.~Liu, S.~Wu, and L.~Wang, ``Deep graph contrastive representation learning,'' \emph{arXiv preprint arXiv:2006.04131}, 2020.

\bibitem{you2020graph}
Y.~You, T.~Chen, Y.~Sui, T.~Chen, Z.~Wang, and Y.~Shen, ``Graph contrastive learning with augmentations,'' \emph{NIPS}, vol.~33, pp. 5812--5823, 2020.

\bibitem{gao2021simcse}
T.~Gao, X.~Yao, and D.~Chen, ``Simcse: Simple contrastive learning of sentence embeddings,'' \emph{arXiv preprint arXiv:2104.08821}, 2021.

\bibitem{kong2019mutual}
L.~Kong, C.~d.~M. d'Autume, W.~Ling, L.~Yu, Z.~Dai, and D.~Yogatama, ``A mutual information maximization perspective of language representation learning,'' \emph{arXiv preprint arXiv:1910.08350}, 2019.

\bibitem{zhao2021graph}
H.~Zhao, X.~Yang, Z.~Wang, E.~Yang, and C.~Deng, ``Graph debiased contrastive learning with joint representation clustering.'' in \emph{IJCAI}, 2021, pp. 3434--3440.

\bibitem{chu2021cuco}
G.~Chu, X.~Wang, C.~Shi, and X.~Jiang, ``Cuco: Graph representation with curriculum contrastive learning.'' in \emph{IJCAI}, 2021, pp. 2300--2306.

\bibitem{xia2022progcl}
J.~Xia, L.~Wu, G.~Wang, J.~Chen, and S.~Z. Li, ``Progcl: Rethinking hard negative mining in graph contrastive learning,'' in \emph{ICML}.\hskip 1em plus 0.5em minus 0.4em\relax PMLR, 2022, pp. 24\,332--24\,346.

\bibitem{zhang2021virtual}
D.~Zhang, W.~Xiao, H.~Zhu, X.~Ma, and A.~O. Arnold, ``Virtual augmentation supported contrastive learning of sentence representations,'' \emph{arXiv preprint arXiv:2110.08552}, 2021.

\bibitem{wang2022sncse}
H.~Wang, Y.~Li, Z.~Huang, Y.~Dou, L.~Kong, and J.~Shao, ``Sncse: Contrastive learning for unsupervised sentence embedding with soft negative samples,'' \emph{arXiv preprint arXiv:2201.05979}, 2022.

\bibitem{xiong2020approximate}
L.~Xiong, C.~Xiong, Y.~Li, K.-F. Tang, J.~Liu, P.~Bennett, J.~Ahmed, and A.~Overwijk, ``Approximate nearest neighbor negative contrastive learning for dense text retrieval,'' \emph{arXiv preprint arXiv:2007.00808}, 2020.

\bibitem{rowley1995human}
H.~Rowley, S.~Baluja, and T.~Kanade, ``Human face detection in visual scenes,'' \emph{NIPS}, vol.~8, 1995.

\bibitem{sung1998example}
K.-K. Sung and T.~Poggio, ``Example-based learning for view-based human face detection,'' \emph{TPAMI}, vol.~20, no.~1, pp. 39--51, 1998.

\bibitem{tran2019improving}
V.-A. Tran, R.~Hennequin, J.~Royo-Letelier, and M.~Moussallam, ``Improving collaborative metric learning with efficient negative sampling,'' in \emph{SIGIR}, 2019, pp. 1201--1204.

\bibitem{simo2015discriminative}
E.~Simo-Serra, E.~Trulls, L.~Ferraz, I.~Kokkinos, P.~Fua, and F.~Moreno-Noguer, ``Discriminative learning of deep convolutional feature point descriptors,'' in \emph{ICCV}, 2015, pp. 118--126.

\bibitem{loshchilov2015online}
I.~Loshchilov and F.~Hutter, ``Online batch selection for faster training of neural networks,'' \emph{arXiv preprint arXiv:1511.06343}, 2015.

\bibitem{wang2015unsupervised}
X.~Wang and A.~Gupta, ``Unsupervised learning of visual representations using videos,'' in \emph{ICCV}, 2015, pp. 2794--2802.

\bibitem{weston2011wsabie}
J.~\vspace{0mm} Weston, S.~Bengio, and N.~Usunier, ``Wsabie: Scaling up to large vocabulary image annotation,'' in \emph{IJCAI}, 2011.

\bibitem{zhao2015improving}
T.~Zhao, J.~McAuley, and I.~King, ``Improving latent factor models via personalized feature projection for one class recommendation,'' in \emph{CIKM}, 2015, pp. 821--830.

\bibitem{guo2018approximating}
G.~Guo, S.~Ouyang, F.~Yuan, and X.~Wang, ``Approximating word ranking and negative sampling for word embedding.''\hskip 1em plus 0.5em minus 0.4em\relax IJCAI, 2018.

\bibitem{faghri2017vse++}
F.~Faghri, D.~J. Fleet, J.~R. Kiros, and S.~Fidler, ``Vse++: Improving visual-semantic embeddings with hard negatives,'' \emph{arXiv preprint arXiv:1707.05612}, 2017.

\bibitem{guo2018vse}
G.~Guo, S.~Zhai, F.~Yuan, Y.~Liu, and X.~Wang, ``Vse-ens: Visual-semantic embeddings with efficient negative sampling,'' in \emph{AAAI}, vol.~32, no.~1, 2018.

\bibitem{li2019sampling}
J.~Li, C.~Tao, W.~Wu, Y.~Feng, D.~Zhao, and R.~Yan, ``Sampling matters! an empirical study of negative sampling strategies for learning of matching models in retrieval-based dialogue systems,'' in \emph{EMNLP-IJCNLP}, 2019, pp. 1291--1296.

\bibitem{zhang2017mixup}
H.~Zhang, M.~Cisse, Y.~N. Dauphin, and D.~Lopez-Paz, ``mixup: Beyond empirical risk minimization,'' \emph{arXiv preprint arXiv:1710.09412}, 2017.

\bibitem{chen2021incremental}
T.-S. Chen, W.-C. Hung, H.-Y. Tseng, S.-Y. Chien, and M.-H. Yang, ``Incremental false negative detection for contrastive learning,'' \emph{arXiv preprint arXiv:2106.03719}, 2021.

\bibitem{yang2022trading}
C.~Yang, Q.~Wu, J.~Jin, X.~Gao, J.~Pan, and G.~Chen, ``Trading hard negatives and true negatives: A debiased contrastive collaborative filtering approach,'' \emph{arXiv preprint arXiv:2204.11752}, 2022.

\bibitem{williams1992simple}
R.~J. Williams, ``Simple statistical gradient-following algorithms for connectionist reinforcement learning,'' \emph{Machine learning}, vol.~8, no.~3, pp. 229--256, 1992.

\bibitem{chen2021novelty}
C.~Chen, Y.~Xie, S.~Lin, R.~Qiao, J.~Zhou, X.~Tan, Y.~Zhang, and L.~Ma, ``Novelty detection via contrastive learning with negative data augmentation,'' \emph{arXiv preprint arXiv:2106.09958}, 2021.

\bibitem{wang2021samwalker++}
C.~Wang, J.~Chen, S.~Zhou, Q.~Shi, Y.~Feng, and C.~Chen, ``Samwalker++: recommendation with informative sampling strategy,'' \emph{TKDE}, 2021.

\bibitem{fang2020cert}
H.~Fang, S.~Wang, M.~Zhou, J.~Ding, and P.~Xie, ``Cert: Contrastive self-supervised learning for language understanding,'' \emph{arXiv preprint arXiv:2005.12766}, 2020.

\bibitem{ye2019unsupervised}
M.~Ye, X.~Zhang, P.~C. Yuen, and S.-F. Chang, ``Unsupervised embedding learning via invariant and spreading instance feature,'' in \emph{CVPR}, 2019, pp. 6210--6219.

\bibitem{yi2019sampling}
X.~Yi, J.~Yang, L.~Hong, D.~Z. Cheng, L.~Heldt, A.~Kumthekar, Z.~Zhao, L.~Wei, and E.~Chi, ``Sampling-bias-corrected neural modeling for large corpus item recommendations,'' in \emph{RecSys}, 2019, pp. 269--277.

\bibitem{karpukhin2020dense}
V.~Karpukhin, B.~O{\u{g}}uz, S.~Min, P.~Lewis, L.~Wu, S.~Edunov, D.~Chen, and W.-t. Yih, ``Dense passage retrieval for open-domain question answering,'' \emph{arXiv preprint arXiv:2004.04906}, 2020.

\bibitem{qu2020rocketqa}
Y.~Qu, Y.~Ding, J.~Liu, K.~Liu, R.~Ren, W.~X. Zhao, D.~Dong, H.~Wu, and H.~Wang, ``Rocketqa: An optimized training approach to dense passage retrieval for open-domain question answering,'' \emph{arXiv preprint arXiv:2010.08191}, 2020.

\bibitem{zhu2021graph}
Y.~Zhu, Y.~Xu, F.~Yu, Q.~Liu, S.~Wu, and L.~Wang, ``Graph contrastive learning with adaptive augmentation,'' in \emph{WWW}, 2021, pp. 2069--2080.

\bibitem{zhou2022debiased}
K.~Zhou, B.~Zhang, W.~X. Zhao, and J.-R. Wen, ``Debiased contrastive learning of unsupervised sentence representations,'' \emph{arXiv preprint arXiv:2205.00656}, 2022.

\bibitem{huynh2022boosting}
T.~Huynh, S.~Kornblith, M.~R. Walter, M.~Maire, and M.~Khademi, ``Boosting contrastive self-supervised learning with false negative cancellation,'' in \emph{Proceedings of the IEEE/CVF Winter Conference on Applications of Computer Vision}, 2022, pp. 2785--2795.

\bibitem{zhan2021optimizing}
J.~Zhan, J.~Mao, Y.~Liu, J.~Guo, M.~Zhang, and S.~Ma, ``Optimizing dense retrieval model training with hard negatives,'' in \emph{SIGIR}, 2021, pp. 1503--1512.

\bibitem{barkan2016item2vec}
O.~Barkan and N.~Koenigstein, ``Item2vec: neural item embedding for collaborative filtering,'' in \emph{MLSP}.\hskip 1em plus 0.5em minus 0.4em\relax IEEE, 2016, pp. 1--6.

\bibitem{yu2017selection}
H.-F. Yu, M.~Bilenko, and C.-J. Lin, ``Selection of negative samples for one-class matrix factorization,'' in \emph{ICDM}.\hskip 1em plus 0.5em minus 0.4em\relax SIAM, 2017, pp. 363--371.

\bibitem{lian2020personalized}
D.~Lian, Q.~Liu, and E.~Chen, ``Personalized ranking with importance sampling,'' in \emph{Proceedings of The Web Conference 2020}, 2020, pp. 1093--1103.

\bibitem{jin2020sampling}
B.~Jin, D.~Lian, Z.~Liu, Q.~Liu, J.~Ma, X.~Xie, and E.~Chen, ``Sampling-decomposable generative adversarial recommender,'' \emph{NIPS}, vol.~33, pp. 22\,629--22\,639, 2020.

\bibitem{berg2017graph}
R.~v.~d. Berg, T.~N. Kipf, and M.~Welling, ``Graph convolutional matrix completion,'' \emph{arXiv preprint arXiv:1706.02263}, 2017.

\bibitem{wu2019session}
S.~Wu, Y.~Tang, Y.~Zhu, L.~Wang, X.~Xie, and T.~Tan, ``Session-based recommendation with graph neural networks,'' in \emph{AAAI}, vol.~33, no.~01, 2019, pp. 346--353.

\bibitem{gao2018self}
H.~Gao and H.~Huang, ``Self-paced network embedding,'' in \emph{KDD}, 2018, pp. 1406--1415.

\bibitem{armandpour2019robust}
M.~Armandpour, P.~Ding, J.~Huang, and X.~Hu, ``Robust negative sampling for network embedding,'' in \emph{AAAI}, vol.~33, no.~01, 2019, pp. 3191--3198.

\bibitem{zhu2022structure}
Y.~Zhu, Y.~Xu, H.~Cui, C.~Yang, Q.~Liu, and S.~Wu, ``Structure-enhanced heterogeneous graph contrastive learning,'' in \emph{SDM}.\hskip 1em plus 0.5em minus 0.4em\relax SIAM, 2022, pp. 82--90.

\bibitem{kanojia2017enhancing}
V.~Kanojia, H.~Maeda, R.~Togashi, and S.~Fujita, ``Enhancing knowledge graph embedding with probabilistic negative sampling,'' in \emph{WWW}, 2017, pp. 801--802.

\bibitem{kotnis2017analysis}
B.~Kotnis and V.~Nastase, ``Analysis of the impact of negative sampling on link prediction in knowledge graphs,'' \emph{arXiv preprint arXiv:1708.06816}, 2017.

\bibitem{sun2019rotate}
Z.~Sun, Z.-H. Deng, J.-Y. Nie, and J.~Tang, ``Rotate: Knowledge graph embedding by relational rotation in complex space,'' \emph{arXiv preprint arXiv:1902.10197}, 2019.

\bibitem{shan2018confidence}
Y.~Shan, C.~Bu, X.~Liu, S.~Ji, and L.~Li, ``Confidence-aware negative sampling method for noisy knowledge graph embedding,'' in \emph{ICBK}.\hskip 1em plus 0.5em minus 0.4em\relax IEEE, 2018, pp. 33--40.

\bibitem{islam2021simple}
M.~K. Islam, S.~Aridhi, and M.~Sma{\"\i}l-Tabbone, ``Simple negative sampling for link prediction in knowledge graphs,'' in \emph{International Conference on Complex Networks and Their Applications}.\hskip 1em plus 0.5em minus 0.4em\relax Springer, 2021, pp. 549--562.

\bibitem{qin2019knowledge}
S.~Qin, G.~Rao, C.~Bin, L.~Chang, T.~Gu, and W.~Xuan, ``Knowledge graph embedding based on adaptive negative sampling,'' in \emph{International Conference of Pioneering Computer Scientists, Engineers and Educators}.\hskip 1em plus 0.5em minus 0.4em\relax Springer, 2019, pp. 551--563.

\bibitem{goldberg2014word2vec}
Y.~Goldberg and O.~Levy, ``word2vec explained: deriving mikolov et al.'s negative-sampling word-embedding method,'' \emph{arXiv preprint arXiv:1402.3722}, 2014.

\bibitem{li2020empirical}
Y.~Li, L.~Liu, and S.~Shi, ``Empirical analysis of unlabeled entity problem in named entity recognition,'' \emph{arXiv preprint arXiv:2012.05426}, 2020.

\bibitem{li2022rethinking}
Y.~\vspace{0mm} Li, L.~Liu, and S.~Shi, ``Rethinking negative sampling for handling missing entity annotations,'' in \emph{ACL}, 2022, pp. 7188--7197.

\bibitem{guo2018effective}
M.~Guo, Q.~Shen, Y.~Yang, H.~Ge, D.~Cer, G.~H. Abrego, K.~Stevens, N.~Constant, Y.-H. Sung, B.~Strope \emph{et~al.}, ``Effective parallel corpus mining using bilingual sentence embeddings,'' \emph{arXiv preprint arXiv:1807.11906}, 2018.

\bibitem{wu2020clear}
Z.~Wu, S.~Wang, J.~Gu, M.~Khabsa, F.~Sun, and H.~Ma, ``Clear: Contrastive learning for sentence representation,'' \emph{arXiv preprint arXiv:2012.15466}, 2020.

\bibitem{wang2021cline}
D.~Wang, N.~Ding, P.~Li, and H.-T. Zheng, ``Cline: Contrastive learning with semantic negative examples for natural language understanding,'' \emph{arXiv preprint arXiv:2107.00440}, 2021.

\bibitem{zhang2022unsupervised}
Y.~Zhang, R.~Zhang, S.~Mensah, X.~Liu, and Y.~Mao, ``Unsupervised sentence representation via contrastive learning with mixing negatives,'' 2022.

\bibitem{gillick2019learning}
D.~Gillick, S.~Kulkarni, L.~Lansing, A.~Presta, J.~Baldridge, E.~Ie, and D.~Garcia-Olano, ``Learning dense representations for entity retrieval,'' \emph{arXiv preprint arXiv:1909.10506}, 2019.

\bibitem{huang2020embedding}
J.-T. Huang, A.~Sharma, S.~Sun, L.~Xia, D.~Zhang, P.~Pronin, J.~Padmanabhan, G.~Ottaviano, and L.~Yang, ``Embedding-based retrieval in facebook search,'' in \emph{KDD}, 2020, pp. 2553--2561.

\bibitem{oord2018representation}
A.~v.~d. Oord, Y.~Li, and O.~Vinyals, ``Representation learning with contrastive predictive coding,'' \emph{arXiv preprint arXiv:1807.03748}, 2018.

\bibitem{tian2020contrastive}
Y.~Tian, D.~Krishnan, and P.~Isola, ``Contrastive multiview coding,'' in \emph{ECCV}.\hskip 1em plus 0.5em minus 0.4em\relax Springer, 2020, pp. 776--794.

\bibitem{misra2020self}
I.~Misra and L.~v.~d. Maaten, ``Self-supervised learning of pretext-invariant representations,'' in \emph{CVPR}, 2020, pp. 6707--6717.

\bibitem{wang2021understanding}
F.~Wang and H.~Liu, ``Understanding the behaviour of contrastive loss,'' in \emph{CVPR}, 2021, pp. 2495--2504.

\bibitem{ge2021robust}
S.~Ge, S.~Mishra, C.-L. Li, H.~Wang, and D.~Jacobs, ``Robust contrastive learning using negative samples with diminished semantics,'' \emph{NIPS}, vol.~34, pp. 27\,356--27\,368, 2021.

\bibitem{sermanet2016unsupervised}
P.~Sermanet, K.~Xu, and S.~Levine, ``Unsupervised perceptual rewards for imitation learning,'' \emph{arXiv preprint arXiv:1612.06699}, 2016.

\bibitem{sermanet2018time}
P.~Sermanet, C.~Lynch, Y.~Chebotar, J.~Hsu, E.~Jang, S.~Schaal, S.~Levine, and G.~Brain, ``Time-contrastive networks: Self-supervised learning from video,'' in \emph{2018 IEEE international conference on robotics and automation (ICRA)}.\hskip 1em plus 0.5em minus 0.4em\relax IEEE, 2018, pp. 1134--1141.

\bibitem{nair2022r3m}
S.~Nair, A.~Rajeswaran, V.~Kumar, C.~Finn, and A.~Gupta, ``R3m: A universal visual representation for robot manipulation,'' \emph{arXiv preprint arXiv:2203.12601}, 2022.

\bibitem{mao2019metric}
C.~Mao, Z.~Zhong, J.~Yang, C.~Vondrick, and B.~Ray, ``Metric learning for adversarial robustness,'' \emph{NIPS}, vol.~32, 2019.

\bibitem{iscen2018mining}
A.~Iscen, G.~Tolias, Y.~Avrithis, and O.~Chum, ``Mining on manifolds: Metric learning without labels,'' in \emph{CVPR}, 2018, pp. 7642--7651.

\bibitem{yuan2017hard}
Y.~Yuan, K.~Yang, and C.~Zhang, ``Hard-aware deeply cascaded embedding,'' in \emph{ICCV}, 2017, pp. 814--823.

\bibitem{xin2018batch}
X.~Xin, Y.~Fajie, H.~Xiangnan, and J.~Joemon, ``Batch is not heavy: Learning word embeddings from all samples,'' in \emph{ACL}, vol.~8, 2018, pp. 107--132.

\bibitem{chen2019efficient}
C.~Chen, M.~Zhang, C.~Wang, W.~Ma, M.~Li, Y.~Liu, and S.~Ma, ``An efficient adaptive transfer neural network for social-aware recommendation,'' in \emph{SIGIR}, 2019, pp. 225--234.

\bibitem{chen2020efficient}
C.~Chen, M.~Zhang, Y.~Zhang, Y.~Liu, and S.~Ma, ``Efficient neural matrix factorization without sampling for recommendation,'' \emph{TOIS}, vol.~38, no.~2, pp. 1--28, 2020.

\bibitem{chen2020jointly}
C.~\vspace{0mm} Chen, M.~Zhang, W.~Ma, Y.~Liu, and S.~Ma, ``Jointly non-sampling learning for knowledge graph enhanced recommendation,'' in \emph{SIGIR}, 2020, pp. 189--198.

\bibitem{caron2020unsupervised}
M.~Caron, I.~Misra, J.~Mairal, P.~Goyal, P.~Bojanowski, and A.~Joulin, ``Unsupervised learning of visual features by contrasting cluster assignments,'' \emph{NIPS}, vol.~33, pp. 9912--9924, 2020.

\bibitem{grill2020bootstrap}
J.-B. Grill, F.~Strub, F.~Altch{\'e}, C.~Tallec, P.~Richemond, E.~Buchatskaya, C.~Doersch, B.~Avila~Pires, Z.~Guo, M.~Gheshlaghi~Azar \emph{et~al.}, ``Bootstrap your own latent-a new approach to self-supervised learning,'' \emph{NIPS}, vol.~33, pp. 21\,271--21\,284, 2020.

\bibitem{chen2021exploring}
X.~Chen and K.~He, ``Exploring simple siamese representation learning,'' in \emph{CVPR}, 2021, pp. 15\,750--15\,758.

\bibitem{thakoor2021large}
S.~Thakoor, C.~Tallec, M.~G. Azar, M.~Azabou, E.~L. Dyer, R.~Munos, P.~Veli{\v{c}}kovi{\'c}, and M.~Valko, ``Large-scale representation learning on graphs via bootstrapping,'' \emph{arXiv preprint arXiv:2102.06514}, 2021.

\bibitem{he2022masked}
K.~He, X.~Chen, S.~Xie, Y.~Li, P.~Doll{\'a}r, and R.~Girshick, ``Masked autoencoders are scalable vision learners,'' in \emph{CVPR}, 2022, pp. 16\,000--16\,009.

\bibitem{hou2022graphmae}
Z.~Hou, X.~Liu, Y.~Dong, C.~Wang, J.~Tang \emph{et~al.}, ``Graphmae: Self-supervised masked graph autoencoders,'' \emph{arXiv preprint arXiv:2205.10803}, 2022.

\bibitem{arora2019theoretical}
S.~Arora, H.~Khandeparkar, M.~Khodak, O.~Plevrakis, and N.~Saunshi, ``A theoretical analysis of contrastive unsupervised representation learning,'' \emph{arXiv preprint arXiv:1902.09229}, 2019.

\bibitem{wu2021rethinking}
C.~Wu, F.~Wu, and Y.~Huang, ``Rethinking infonce: How many negative samples do you need?'' \emph{arXiv preprint arXiv:2105.13003}, 2021.

\bibitem{ash2021investigating}
J.~T. Ash, S.~Goel, A.~Krishnamurthy, and D.~Misra, ``Investigating the role of negatives in contrastive representation learning,'' \emph{arXiv preprint arXiv:2106.09943}, 2021.

\bibitem{nozawa2021understanding}
K.~Nozawa and I.~Sato, ``Understanding negative samples in instance discriminative self-supervised representation learning,'' \emph{NIPS}, vol.~34, pp. 5784--5797, 2021.

\bibitem{awasthi2022more}
P.~Awasthi, N.~Dikkala, and P.~Kamath, ``Do more negative samples necessarily hurt in contrastive learning?'' \emph{arXiv preprint arXiv:2205.01789}, 2022.

\bibitem{sohn2016improved}
K.~Sohn, ``Improved deep metric learning with multi-class n-pair loss objective,'' \emph{NIPS}, vol.~29, 2016.

\end{thebibliography}
\vspace{-1.8cm}
\begin{IEEEbiography}[{\includegraphics[width=0.8in,height=1in,clip,keepaspectratio]{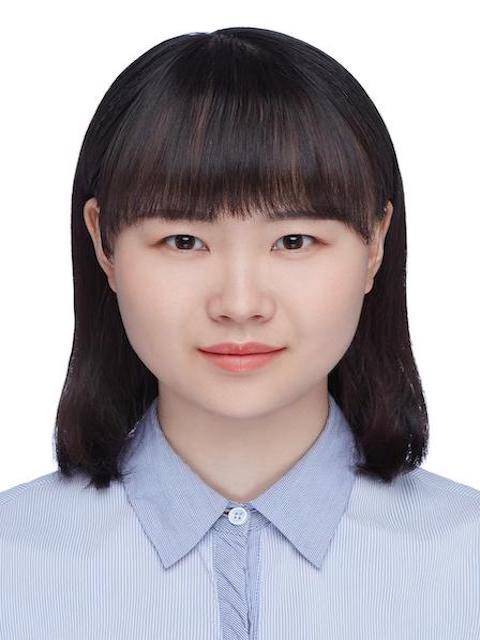}}]{Zhen Yang} is a PhD candidate in the Department of Computer Science and Technology, Tsinghua University. She got her master’s degree from the Institute of Microelectronics, Tsinghua University. Her research interests include graph representation learning, graph-based recommendation, contrastive learning, and large language models (LLMs). She has published some papers on top conferences and journals, such as KDD, WWW, AAAI, ICCV and TKDE.
\end{IEEEbiography}

\vspace{-1.4cm}

\begin{IEEEbiography}[{\includegraphics[width=1.0in,height=1in,clip,keepaspectratio]{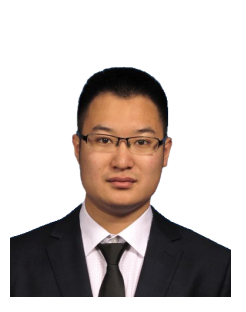}}]{Ming Ding} obtained his Ph.D. degree in 2023 from the Department of Computer Science and Technology, Tsinghua University. He is currently working at Zhipu AI. His research interests include graph learning, natural language processing and cognitive artificial intelligence. He has published many papers on top conferences, such as NeurIPS, KDD, ACL, IJCAI, etc. 
\end{IEEEbiography}

\vspace{-1.4cm}

\begin{IEEEbiography}[{\includegraphics[width=1.3in,height=1in,clip,keepaspectratio]{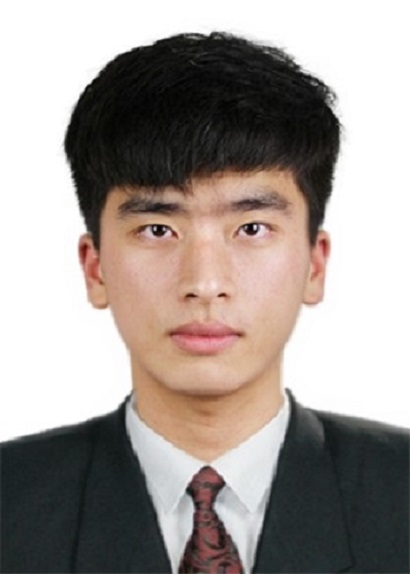}}]{Tinglin Huang} Tinglin Huang is a PhD candidate in the Department of Computer Science, Yale University. He got his master's degree from Zhejiang University. His research interests include computational biology, geometric deep learning, and graph learning.
He has published some papers at top conferences, such as NeurIPS, KDD, WWW, etc.
\end{IEEEbiography}

\vspace{-1.4cm}

\begin{IEEEbiography}[{\includegraphics[width=1.3in,height=1in,clip,keepaspectratio]{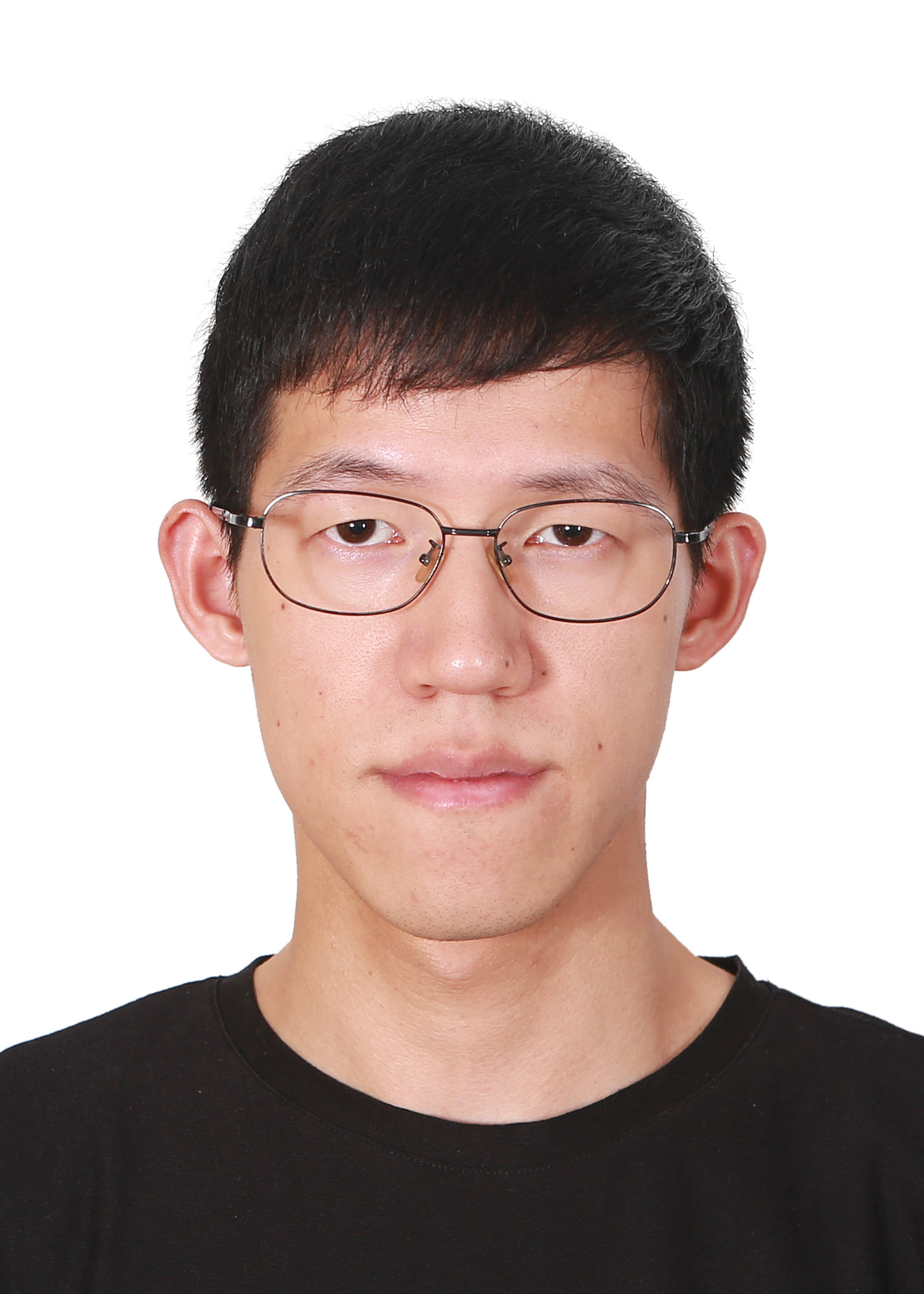}}]{Yukuo Cen} obtained his Ph.D. degree in 2023 from the Department of Computer Science and Technology, Tsinghua University. He also got his bachelor's degree from Tsinghua University. He is currently working at Zhipu AI. He has published several papers on the top international conferences and journals, including KDD, WWW, and TKDE, on graph representation learning and recommender systems.
\end{IEEEbiography}

\vspace{-1.4cm}

\begin{IEEEbiography}[{\includegraphics[width=1.3in,height=1in,clip,keepaspectratio]{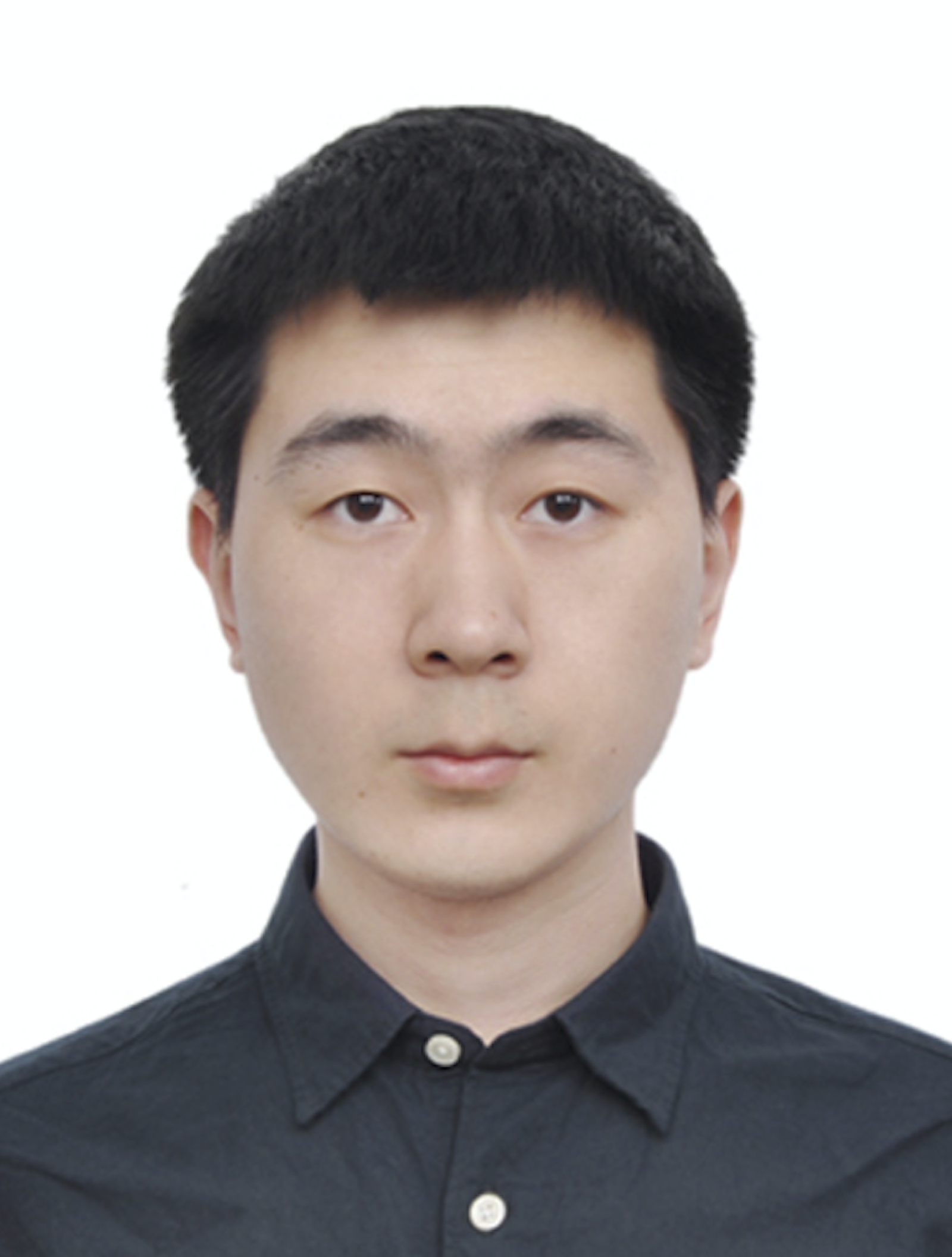}}]{Junshuai Song} received his Ph.D. degree from Peking University in 2021. He has published papers in top conferences and journals, including TKDE, ICDE, ACL, and AAAI on graph mining, network representation learning, and recommender systems. 
\end{IEEEbiography}

\vspace{-1.5cm}

\begin{IEEEbiography}[{\includegraphics[width=1.3in,height=1in,clip,keepaspectratio]{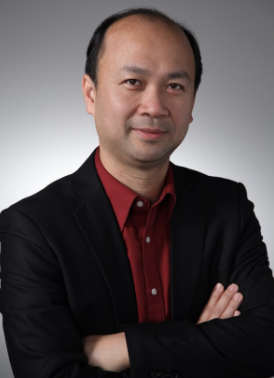}}]{Bin Xu} is a Professor of Computer Science at Tsinghua University. He became ACM Professional member in 2009 and IEEE member in 2007. His research interest is Knowledge Gragh and Large Language Models (LLMs). He is core member to develop GLM family of LLMs.
\end{IEEEbiography}

\vspace{-1.4cm}

\begin{IEEEbiography}[{\includegraphics[width=1.3in,height=1in,clip,keepaspectratio]{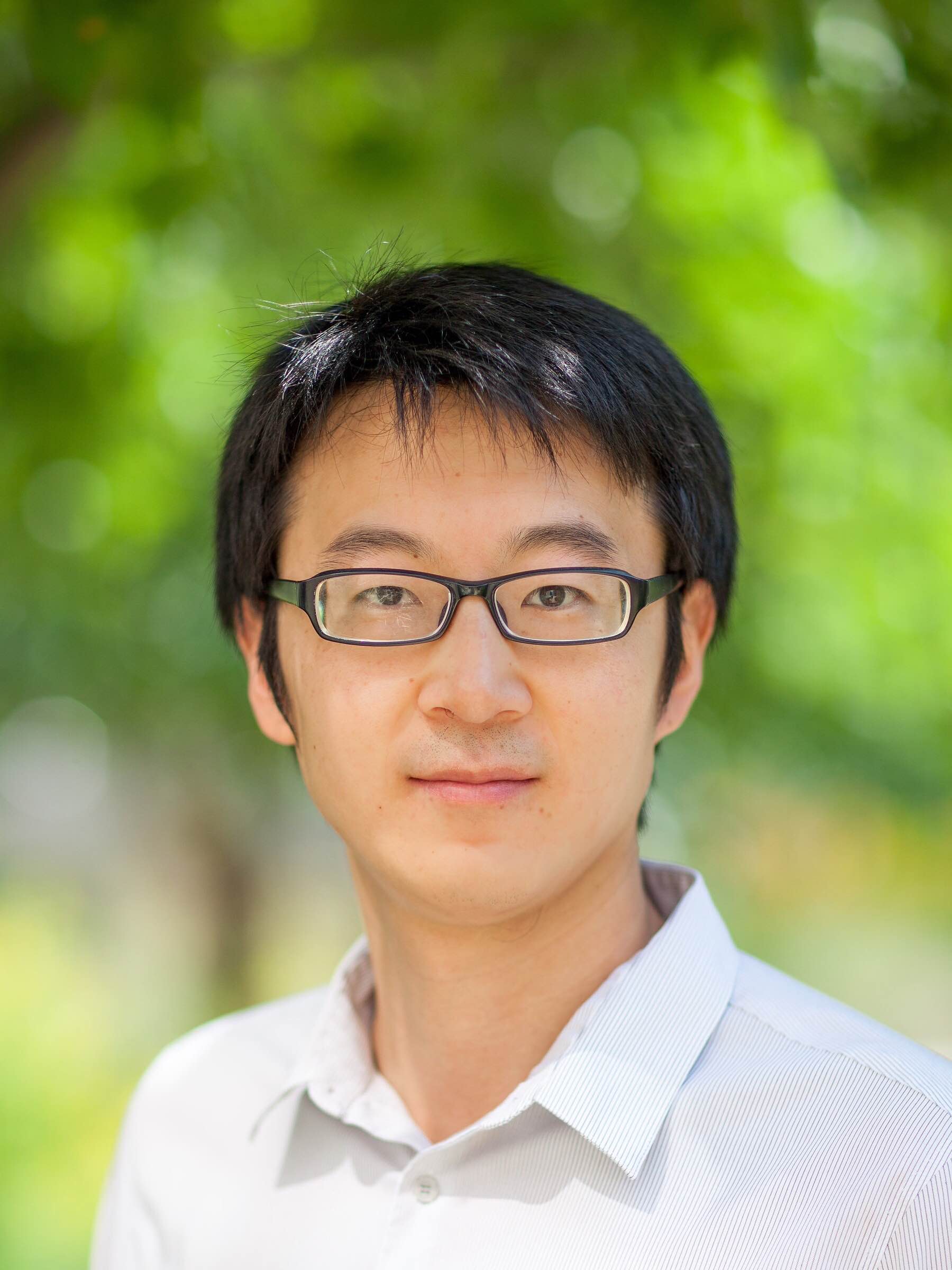}}]{Yuxiao Dong} 
is an associate professor of computer science at Tsinghua University. 
  His research focuses on data mining, graph machine learning, and foundation models. 
  Together with collaborators, his recent research includes GNNs (hetero. graph transformer, GRAND, OGB), network embedding (metapath2vec, NetMF, NetSMF, SketchNE), graph pre-training (GraphMAE, GPT-GNN, GCC), and LLMs (GLM-130B, ChatGLM, CodeGeeX). 
  He received the 
  2022 SIGKDD Rising Star Award. 
\end{IEEEbiography}

\vspace{-1.2cm}

\begin{IEEEbiography}[{\includegraphics[width=0.8in,height=1in,clip,keepaspectratio]{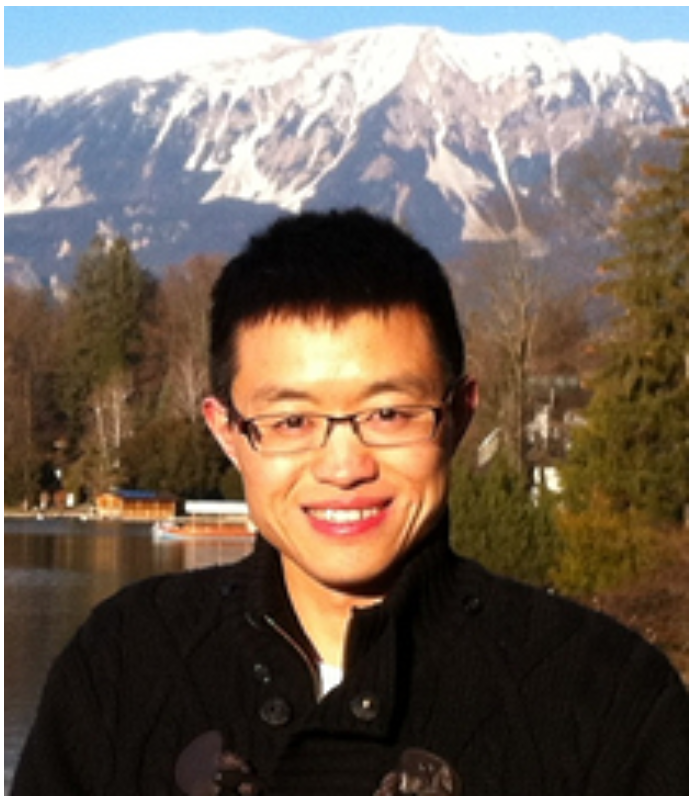}}]{Jie Tang} is a WeBank Chair Professor of Computer Science at Tsinghua University. He is a Fellow of the ACM, a Fellow of AAAI, and a Fellow of IEEE. His research interest is in artificial general intelligence (AGI). His research has received the SIGKDD Test-of-Time Award (10-year Best Paper). He also received the SIGKDD Service Award. Recently, he puts all his efforts into Large Language Models (LLMs). 
He invented AMiner.org and developed the GLM family of LLMs, such as ChatGLM, CogVLM, CodeGeeX, and CogView. 
He served as general chair of WWW'23, program chair of WWW'22 and WSDM'15, and EiC of IEEE Trans. on Big Data and AI Open J.
\end{IEEEbiography}

\end{document}